\documentclass[twoside,11pt]{article}

\usepackage{jmlr2e}

\usepackage{amsmath}
\usepackage{amssymb}
\usepackage{color}
\usepackage{colortbl}
\usepackage{tabularx}
\usepackage{subfigure}
\usepackage{arydshln}
\usepackage{url}
\usepackage{multirow}
\usepackage{epstopdf}
\usepackage{pbox}
\usepackage{footnote}

\newcommand{\diag}{\operatornamewithlimits{diag}}
\newcommand{\sign}{\operatornamewithlimits{sign}}

\newcommand{\bx}{\mathbf{x}}
\newcommand{\by}{\mathbf{y}}
\newcommand{\bz}{\mathbf{z}}

\newcommand{\bw}{\mathbf{w}}

\newcommand{\bX}{\mathbf{X}}
\newcommand{\bH}{\mathbf{H}}
\newcommand{\noisyx}{\tilde{x}}
\newcommand{\noisybx}{\tilde{\mathbf{x}}}

\definecolor{gray}{gray}{0.7}
\firstpageno{1}

\begin{document}

\title{Marginalizing Corrupted Features}

\author{\name Laurens van der Maaten \email lvdmaaten@gmail.com\\
       \addr Pattern Recognition and Bioinformatics Group\\
       Delft University of Technology\\
       Mekelweg 4, 2628 CD Delft, The Netherlands
       \AND
       \name Minmin Chen\footnote{Most of this work was performed while the author was at Washington University in St. Louis.} \email m.chen@criteo.com\\
       \addr Criteo\\
       Palo Alto, CA 94301, USA
       \AND
       \name Stephen Tyree \email swtyree@wustl.edu\\
       \name Kilian Q. Weinberger \email kilian@wustl.edu\\
       \addr Department of Computer Science and Engineering\\
       Washington University in St. Louis\\
       St. Louis, MO 63130, USA}

\editor{}

\maketitle

\begin{abstract}
The goal of machine learning is to develop predictors that generalize well to test data. Ideally, this is achieved by training on an almost infinitely large training data set that captures all variations in the data distribution. In practical learning settings, however, we do not have infinite data and our predictors may overfit. Overfitting may be combatted, for example, by adding a regularizer to the training objective or by defining a prior over the model parameters and performing Bayesian inference. In this paper, we propose a third, alternative approach to combat overfitting: we extend the training set with infinitely many artificial training examples that are obtained by corrupting the original training data. We show that this approach is practical and efficient for a range of predictors and corruption models. Our approach, called \emph{marginalized corrupted features} (MCF), trains robust predictors by minimizing the expected value of the loss function under the corruption model. We show empirically on a variety of data sets that MCF classifiers can be trained efficiently, may generalize substantially better to test data, and are also more robust to feature deletion at test time.  
\end{abstract}


\section{Introduction}
\label{Introduction}  

Dealing with overfitting is one of the key problems one encounters when training machine-learning models. When a learner overfits, it performs substantially better on its training data than on held-out test data. An overfitted learner relates the label or target value with patterns within the noise instead of within the data distribution. 
Three approaches are commonly used to combat overfitting: 1. \emph{early-stopping} techniques monitor the performance of the model on a validation set during training, and stop the learning as soon as the validation performance deteriorates; 2. \emph{regularization} techniques encourage the learning to find ``simple'' models by penalizing ``complex'' models, {\it e.g.}, models with large parameter values; and 3. \emph{Bayesian} techniques define a prior distribution over models that favor simple models, and perform predictions by averaging over the model posterior. Regularization and Bayesian techniques are intimately related because regularization techniques can often be viewed as introducing a prior over models and performing maximum-likelihood estimation. 

In this paper, we propose a new alternative to counter overfitting. Instead of requiring the user to define prior distributions or regularizers over \emph{model parameters}, which can be very counter intuitive, we focus on corruptions of the \emph{data}. 
Our approach is based on the observation that overfitting would completely disappear if we were to train our models on infinite data drawn from the data distribution ${\cal P}$. In such a hypothetical scenario, it is impossible to overfit models; and even high-variance models, such as the nearest neighbor classifier~\citep{cover1967nearest}, become close to optimal ({\it viz.} the error of a nearest neighbor classifier is twice the Bayes error). Unfortunately, a learning scenario in which we only obtain a finite training set is more realistic; here, some variations in the data distribution will not be captured and the learned model performs worse at test time than during training. 

In many learning scenarios, we may however have some additional knowledge about the data distribution: we might know that certain \emph{corruptions} of data instances do not affect their conditional label distributions. As an example, deleting a few words in a text document rarely changes its topic. With this prior knowledge, we can corrupt existing data to generate new artificial instances that resemble those sampled from the actual data distribution. 
In fact, we will \emph{corrupt} the existing \emph{finite} training examples with a fixed corrupting distribution to construct an \emph{infinite} training set. We show that for a wide range of learning models and noise distributions, it is practical to train models on such an infinite, augmented training set. We refer to the resulting framework as learning with \emph{marginalized corrupted features} (MCF). So instead of approximating the exact statistics of ${\cal P}$ with \emph{finite} data, MCF learns from a slightly modified data distribution ${\cal P}'$ with \emph{infinite} data. 

\citet{burges1997improving} explicitly augment the training set with additional examples that are corrupted through similar transformations. Although the simplicity of such an approach is appealing, it lacks elegance and the computational cost of processing the additional corrupted training examples is prohibitive for most real-world problems. In contrast, we show that it is efficient to train predictors on an infinite amount of corrupted copies of the training data by marginalizing out the corrupting distribution. In particular, we focus on empirical risk minimization and derive analytical solutions for the expected loss under a large family of corrupting distributions for quadratic and exponential loss functions. This allows us to minimize the expected loss in computational time linear in the number of training examples. For logistic loss functions, which are used in many probabilistic models, we derive practical approximations and upper bounds for the expected value of the loss under the corrupting distribution.

Our augmented data distribution ${\cal P}'$ is constructed using a simple stochastic rule: pick one of the finite training examples uniformly at random and transform it with some pre-defined corrupting distribution. Many corrupting distributions are possible, but in this paper we focus on 1. Gaussian corruption, 2. Poisson corruption, and 3. (unbiased) blankout or dropout corruption (random deletion of features). The Gaussian corruption model is mainly of interest for continuous-valued data sets; special cases of MCF with Gaussian corruption have already been studied in the context of Parzen density estimation \citep{parzen62} and in the context of vicinal risk minimization \citep{chapelle00}. The Poisson corruption model is of interest when the data comprises count vectors, {\it e.g.}, in document classification. Poisson corruption is particularly appealing as it introduces \emph{no additional hyper-parameters} and, in our results, improves the test accuracy on almost all data sets that comprise count data. The blankout corruption model is of interest in data sets with heavy-tailed feature distributions, such as filter responses, and in settings where blankout noise is a known source of variance in the original data distribution ${\cal P}$---possibly unobserved in the training data. This happens, \emph{e.g.}, in document classification from term-frequency vectors: a big portion of each document (especially after stop-word removal) is sampled from the tail of the power-law distribution, for which the blankout noise model is a surprisingly good approximation: it models the common case that some of the words related to the class of the document are missing, \emph{e.g.} because the author used other synonyms. Blankout corruption is also of interest in the ``nightmare at test time'' scenario \citep{globerson06} in which some of the features are deleted during testing, \emph{e.g.}, due to sensors failures or because the feature computation exceeds a time budget, and we wish to make predictions that are robust to this feature deletion. Different from previous work~\citep{bhattacharyya04,globerson06,shivaswamy06,trafalis07}, our setting focuses on random feature removal and not on the (in practice much rarer) worst-case adversarial setting.  

An earlier short paper on this work \citep{vandermaaten13} already introduces the MCF learning framework, but the present paper provides a large amount of additional details, analysis, and experiments. The paper makes the following contributions: 1. we introduce learning with marginalized corrupted features, a framework that trains robust classifiers by marginalizing out all possible feature corruptions from a pre-defined distribution; 2. we derive plug-in solutions for the quadratic, exponential, and logistic loss functions for a range of corrupting distributions, which can be incorporated into learning algorithms out-of-the-box; and 3. on several real-world data sets, we show that training with MCF may lead to more robust classifiers and may substantially decrease generalization errors in various learning settings. Given the simplicity and elegance of MCF, it could become a valuable alternative to the common $l_1$ or $l_2$-norm regularizers. 

The outline of the remainder of this paper is as follows. In Section~\ref{Related work}, we discuss prior work that is related to the work presented in this paper. In Section~\ref{Robust classifiers}, we introduce marginalized corrupted features (MCF) regularization and we derive MCF variants of quadratic and exponential loss. In Section~\ref{Logistic loss}, we present approximations and upper bounds that allow MCF to be used with logistic loss functions. Section~\ref{Experiments} presents an extensive experimental evaluation in which we compare MCF-regularized predictors with standard predictors in various document and image classification problems, in a classification problem from bioinformatics, and in the ``nightmare at test time scenario''. Section~\ref{Discussion} concludes the paper with a discussion of the results presented in the paper as well as directions for future work.

\section{Related work}
\label{Related work}

There are two motivations for training a classifier with MCF: 1. to reduce the effects of overfitting during \emph{training}; and 2. to combat the effect of data corruption during \emph{testing}. Both motivations connect with different lines of prior work, and in this section we discuss both of them. 


\subsection{Corruption during training}
The initial publication by \cite{burges1997improving} has inspired various lines of work that \emph{explicitly} corrupt training data during training. Most prominently, \citet{vincent08,vincent10} propose to randomly blank out features in the training data that is used as input to autoencoders, whilst leaving the desired output (the original training data) unaltered. The resulting denoising autoencoder model is now commonly used as a building block in deep learning \citep{maillet09,glorot11,mesnil11}, and blankout corruption is also increasingly applied on the hidden units of neural networks as a form of regularization \citep{hinton12}. Since autoencoders are non-linear, the marginalization over the corrupting distribution cannot be performed analytically in such models. Linear denoising autoencoders~\citep{chen2012msda} can be viewed as a special case of MCF that aim to minimize the expected value of the reconstruction error under blankout corruption---and which are stacked in multiple layers. \citet{herbrich2004invariant} propose an elegant generalization of SVMs that learns to be invariant to polynomial input transformations via a semi-definite programming formulation. 

In online learning and bandit problems, various studies have shown that it is possible to learn from data that are subject to a (possibly unknown) corrupting distribution \citep{flaxman05,abernethy08,cesabianchi10}. For instance, \citet{cesabianchi11} show that noise in the data does not affect the convergence rate of online learners. A similar regret bound is proven by \citet{rostamizadeh11} for online learning under the presence of randomly missing features (\emph{i.e.} blankout corruption). These studies differ from our work in that they show that (unknown) corruptions in the data do not \emph{impair} learning performance too much. By contrast, our work shows that adding corruptions to the data can actually \emph{improve} learning performance.

\subsection{Corruption during testing}
Several prior studies consider {\it implicit} approaches for data sets that are subject to corruptions during \emph{test time} (this is also known as ``the nightmare at test-time'' scenario). 
Most of these studies propose to minimize the loss under an adversarial worst-case scenario. In particular, \citet{globerson06} propose a minimax-formulation in which the loss is minimized assuming maximum ``damage'' through corruption. 
By contrast, \citet{dekel08} propose a linear-programming formulation that minimizes an approximation to the same quantity for margin-based predictors. Other studies \citep{bhattacharyya04,shivaswamy06,trafalis07,xu09} also use minimax approaches that minimize losses under a worst-case scenario, but they focus on a somewhat different corruption model that adds a uniformly drawn constant, within a fixed interval, to each feature.  \citet{livni12} consider a scenario in which an adversarial chooses the corrupting distribution from the set of all distributions with a pre-specified variance as to increase the expected loss under the corruption as much as possible. \citet{chechik08} propose an algorithm that maximizes the margin in the subspace of the observed features for each training instance to be robust against random feature deletion. \citet{teo2008convex} generalize the worst-case scenario from simple feature corruption to obtain invariances to transformations such as image rotations or translations. Their framework incorporates several prior formulations on learning with invariants as special cases, most prominently, the formulation of \citet{herbrich2004invariant}. 

Such previous work differs from our approach in that it focuses on adversarial scenarios and does not consider the corruption analytically in expectation. Further, it mostly focuses on a very specific corruption model. In particular, existing approaches suffer from three main disadvantages: 1. they are complex and computationally expensive; 2. they minimize the loss of a worst-case scenario that is very unlikely to be encountered in practice; and 3. they do not provide a flexible framework that can be used with a variety of models and corrupting distributions. By contrast, we propose a much simpler approach that scales linearly in the number of training samples, that considers an average-case instead of a worst-case scenario, and that can readily be used with a variety of loss functions and a large family of corruption models.

\section{Learning with Marginalized Corrupted Features (MCF)}
\label{Robust classifiers}

To derive the \emph{marginalized corrupted features} (MCF) framework, we start by defining a corrupting distribution that specifies how training observations $\bx$ are transformed into corrupted versions $\noisybx$. Throughout the paper, we assume that the corrupting distribution factorizes over dimensions and that each individual distribution $P_E$ is a member of the natural exponential family. (We will see later that for MCF with quadratic loss functions, these assumptions may be relaxed.) Specifically, we assume a corrupting distribution of the form:
\begin{equation}
p(\noisybx | \bx)  =  \prod_{d=1}^D P_{E}(\noisyx_d | x_d; \eta_d),\label{eq:noisemodel2}
\end{equation}
where $\eta_d$ represents user-defined hyperparameters of the corrupting distribution on dimension $d$. Corrupting distributions of interest, for $P_{E}$, include: 1. independent salt or ``blankout''  (or dropout) noise in which the $d$-th feature is randomly set to zero with probability $q_d$; 2. bit-swap noise in which the value of the $d$-th bit is randomly swapped with probability $q_d$; 3. independent Gaussian noise on the $d$-th feature with variance $\sigma_d^2$; 4. independent Laplace noise on the $d$-th feature with variance $2 \lambda_d^2$; and 5. independent Poisson corruption in which the $d$-th feature is used as the rate of the Poisson distribution. Note that the definition in~(\ref{eq:noisemodel2}) allows for different features to use arbitrary different corrupting distributions. In the remainder of this paper, we propose an efficient way to train a classifier on corrupted inputs $\noisybx$, sampled from the distribution $p(\noisybx |\bx){\cal P}(\bx)$. There are at least two motivations for doing so: 

\paragraph{1. Nightmare at test-time.}
In the \emph{nightmare at test-time} scenario, introduced by~\cite{globerson06}, there is an expectation that corruption will appear during test-time. Although the training data is sampled from ${\cal P}(\bx)$, the test data will be sampled from the distribution $p(\noisybx |\bx){\cal P}(\bx)$. Training the classifier on the corrupted data corrects this distribution drift. This scenario appears, for example, in the context of search engines. A feature can be reliably collected for the training set, but during testing it is dropped if its computation time exceeds a pre-specified time limit. The frequency of these feature drop-outs can be measured and incorporated into training with the appropriate dropout distribution. Similarly, one can imagine scenarios (\emph{e.g.} in robotics or sensor networks) where features correspond to unreliable sensor readings. 

\paragraph{2. Regularization.} The more common motivation is improved generalization. Here, the corrupting distribution is meant to capture some of the variance inherent in ${\cal P}(\bx)$. 
In this setting, one must guarantee that the classifier is still applicable for the test case, where data is sampled directly from ${\cal P}$. More precisely, it must be \emph{unbiased}. Let $h(\bx)$ be the classification function, we therefore require of the corrupting distribution that 
\begin{equation}
\mathbb{E}[h(\noisybx)]_{p(\noisybx|\bx)}=h(\bx).     \label{eq:unbiased}
\end{equation}
In other words, corrupting an input should not change its expected prediction and a test input obtains the expected prediction that the same input would obtain during training. 
In the linear case, the classifier is parameterized by a vector $\bw$. For regression, the prediction function is defined as $h(\bx)=\bw^\top\bx$ and we require $\mathbb{E}[\noisybx]=\bx$. For classification, it is $h(\bx)=\textrm{sign}(\bw^\top\bx)$, which necessitates the slightly weaker condition $\mathbb{E}[\noisybx]=s(\bx)\bx$ for any bounded $s(\bx)>0$. 
Table~\ref{table:overview1} showcases several biased and unbiased distributions.

\paragraph{}Assume we are provided with a training data set $\mathcal{D}  =  \{(\bx_n ,  y_n) \}_{n=1}^N$ and a loss function $L(\bx, y; \Theta)$, with model parameters $\Theta$. 
A simple approach to approximately 
learn from the distribution $p(\noisybx |\bx){\cal P}(\bx)$ is to follow the spirit 
of~\citet{burges1997improving} and corrupt each training sample $M$ times, following (\ref{eq:noisemodel2}). 
For each $\bx_n\in{\cal D}$, this results in corresponding corrupted observations $\noisybx_{nm}$ (with $m  =  1, \dots, M$) and leads to the construction of a new data set $\tilde{\mathcal{D}}$ of size $|\tilde{\mathcal{D}}| = MN$. This new data set can be used for training in an empirical risk minimization framework by minimizing the surrogate loss function:
\begin{equation}
\mathcal{L}(\tilde{\mathcal{D}}; \Theta) = \frac{1}{N} \sum_{n=1}^N \frac{1}{M} \sum_{m=1}^M L(\noisybx_{nm}, y_n; \Theta),\label{eq:explicitM}
\end{equation}
with $\noisybx_{nm} \sim p(\noisybx_{nm} | \bx_n)$. Such approaches have recently become popular, in particular, in the deep-learning community as a way to regularize deep networks \citep{hinton12}.

Although approaches that explicitly corrupt the training data are effective, they lack elegance and come with high computational costs: the minimization of $\mathcal{L}(\tilde{\mathcal{D}}; \Theta)$ scales linearly in the number of corrupted observations, \emph{i.e.} it scales as $\mathcal{O}(NM)$. It is, however, of interest to consider the limiting case in which $M \rightarrow \infty$. In this case, we can apply the \emph{weak law of large numbers} and rewrite $\frac{1}{M}\sum_{m=1}^ML(\noisybx_m,y_m;\Theta)$ as its expectation \citep[\S 2.10.2]{duda01} to obtain the following surrogate loss:
\begin{equation}
\mathcal{L}(\mathcal{D}; \Theta) = \frac{1}{N} \sum_{n=1}^N \mathbb{E}[L(\noisybx_n, y_n; \Theta)]_{p(\noisybx_n | \bx_n)}.\label{eq:regret}
\end{equation}
Minimizing the expected value of the loss under the corruption model leads to a new approach for training predictors that we refer to as learning with \emph{marginalized corrupted features} (MCF). MCF may lead to algorithms with an $\mathcal{O}(N)$ training complexity in situations in which the expectation in (\ref{eq:regret}) is tractable, which is indeed the case for commonly used loss functions $L(\noisybx_n, y_n; \Theta)$ and corrupting distributions $p(\noisybx | \bx)$.

In the following, we show that for linear predictors that employ a quadratic or exponential loss function, the required expectations under $p(\noisybx | \bx)$ in (\ref{eq:regret}) can be computed analytically for all corrupting distributions in the natural exponential family. For linear predictors with logistic loss we derive practical approximations and upper bounds for the expected loss under $p(\noisybx | \bx)$, which may serve as surrogate loss functions. 

\subsection{Quadratic loss}
\label{Quadratic loss}
Assuming a linear model parametrized by vector $\bw$ and a target variable $y$ (for regression, $y$ is continuous; for binary classification, $y  \in  \{-1, +1\}$), the expected value of the quadratic loss under corrupting distribution $p(\noisybx | \bx)$ is given by:
\begin{align}
\mathcal{L}(\mathcal{D}; \bw) &= \frac{1}{N} \sum_{n=1}^N \mathbb{E}\left[\bigg(\bw^\top \noisybx_n - y_n \bigg)^2\right]_{p(\noisybx_n | \bx_n)}\nonumber\\
&= \frac{1}{N} \sum_{n=1}^N \mathbb{E}\left[\bw^\top \noisybx_n \noisybx_n^\top \bw - 2y_n \bw^\top\noisybx_n + y_n^2 \right]_{p(\noisybx_n | \bx_n)}\nonumber\\
&= \bw^\top \bigg(\frac{1}{N} \sum_{n=1}^N \left[ \mathbb{E}[\noisybx_n] \mathbb{E}[\noisybx_n]^\top + \diag\left(\mathbb{V}[\noisybx_n]\right) \right] \bigg) \bw - 2\bigg(\frac{1}{N} \sum_{n=1}^N y_n \mathbb{E}\left[\noisybx_n \right]\bigg)^\top \bw + 1.\label{eq:final_lls}
\end{align}
Herein, $\mathbb{V}[x]$ denotes the variance of $x$, and all expectations are under $p(\noisybx_n | \bx_n)$. 
Note that irrespective of what corruption model is used, (\ref{eq:final_lls}) is quadratic in $\bw$, and hence convex. The optimal solution $\bw^*$ can be computed with a variation of the ordinary least squares closed-form solution:
\begin{equation}
\bw^* = \bigg(\sum_{n=1}^N \left[ \mathbb{E}[\noisybx_n] \mathbb{E}[\noisybx_n]^\top + \diag\left(\mathbb{V}[\noisybx_n]\right) \right] \bigg)^{-1} \bigg(\sum_{n=1}^N y_n \mathbb{E}\left[\noisybx_n \right]\bigg).\label{eq:optimal_w_lls}
\end{equation}
Hence, to minimize the expected quadratic loss under the corruption model, we only need to compute the mean and variance of the corrupting distribution, which is practical for all exponential-family distributions (as well as for certain distributions outside of the exponential family, such as Student-t distributions with more than two degrees of freedom). Table~\ref{table:overview1} gives an overview of the required mean and variance for six corrupting distributions of interest.

\begin{table*}[t]
    \small
\begin{center}
	\includegraphics[width=\textwidth]{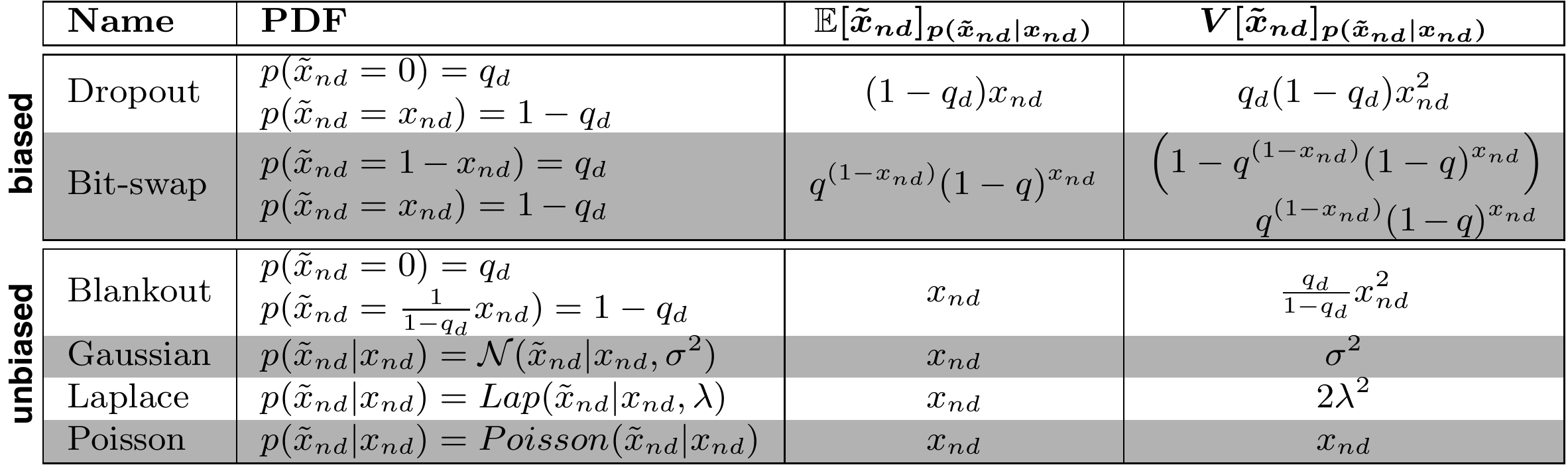}
\caption{The probability density function (PDF), mean, and variance of corrupting distributions of interest. These quantities can be plugged into Eq.~(\ref{eq:final_lls}) to obtain the expected value under the corrupting distribution of the quadratic loss. For disambiguation, we refer to the unbiased version of feature dropout as blankout corruption. }\label{table:overview1}
\end{center}
\end{table*}

\paragraph{Special case.} An interesting setting of MCF with quadratic loss occurs when the corrupting distribution $p(\noisybx | \bx)$ is an isotropic Gaussian distribution with mean $\bx$ and variance $\sigma^2\mathbf{I}$. For such a Gaussian corruption model, we obtain the standard $l_2$-regularized quadratic loss with regularization parameter $\sigma^2$ that is used in ridge regression as special case~\citep{chapelle00}:
\begin{equation}
\mathcal{L}(\mathcal{D}; \bw) = \bw^\top \bigg(\frac{1}{N} \sum_{n=1}^N \bx_n \bx_n^\top\bigg) \bw - 2\bigg(\frac{1}{N} \sum_{n=1}^N y_n \bx_n\bigg)^\top \bw + \sigma^2 \bw^\top \bw + 1.
\end{equation}
Perhaps surprisingly, using MCF with Laplace corruption also leads to ridge regression with the regularization parameter taking value $2 \lambda^2$. Indeed, ridge regression arises for every unbiased corrupting distribution whose variance is not data-dependent. For corrupting distributions whose variance is data-dependent, other regularizers are obtained (see below).


\paragraph{Blankout corruption.} We study the use of blankout corruption in MCF quadratic loss in more detail here to show what kind of regularizers it produces. By plugging the relevant quantities from Table~\ref{table:overview1} into (\ref{eq:optimal_w_lls}), we obtain the following solution for $\bw^*$ with MCF blankout (assuming $\forall d: q_d = q$):
\begin{align}
\bw^* = \left( \sum_{n=1}^N \bx_n \bx_n^\top + q(1 - q) \mathbf{S} \right)^{-1} \left( \sum_{n=1}^N \bx_n y_n\right), \textrm{ where: } \mathbf{S}=\left(\sum_{n=1}^N \bx_n\bx_n^\top\right) \circ \mathbf{I} 
\end{align}
where $\circ$ denotes an element-wise Hadamard product. In words, the matrix $\mathbf{S}$ consists of only the diagonal of the Gram matrix.  
Consequently, MCF blankout corruption has the effect of multiplying the diagonal of the Gram matrix by a factor $(1+q-q^2)$. This makes the MCF regularizer \emph{data-dependent}: the regularization is stronger on dimensions that on average have a larger norm. This may be desirable in situations in which the data dimensions live on different scales as happens, for example, in word-count features.
This is in contrast to the \emph{data-independent} $l_2$-regularizer, that simply adds a fixed value $\lambda$ to the diagonal of the scatter matrix---treating all features alike. Similar data-dependent regularizers have previously been studied by \citet{grave11}.

\paragraph{Poisson corruption.} Using Poisson corruption in MCF quadratic loss leads to:
\begin{align}
\bw^* = \left(\sum_{n=1}^N \bx_n \bx_n^\top + \diag\left(\sum_{n=1}^N \bx_n \right) \right)^{-1} \left(\sum_{n=1}^N \bx_n y_n\right).
\end{align}
Poisson corruption thus leads to a \emph{parameterless, additive, data-dependent} regularizer that adds larger values to the diagonal of the Gram matrix for features that occur more on average. (Note that Poisson corruption can only be used on non-negative data, $\forall n,d: x_{nd} \geq 0$, so the contribution of the regularizer is always non-negative.) The main difference between blankout and Poisson corruption is in how strongly features with large values are regularized; the multiplicative blankout corruption regularizes common features more than the additive Poisson corruption.

\paragraph{Leave-one-out errors.} A nice property of linear models employing quadratic loss is that they allow us to compute the leave-one-out error of the model from only the training predictions $\hat{y}_n$ \citep{allen74}. We show that this property still holds for MCF quadratic losses, irrespective of the corrupting distribution that is used. Defining the target vector $\by = [y_1, \dots, y_N]^\top$ and the data matrix $\bX = [\bx_1, \dots, \bx_N]$, we can rewrite the prediction vector $\hat{\by} = [\hat{y}_1, \dots, \hat{y}_N]$ as:
\begin{equation}
\hat{\by} = \bX^\top \bw^* = \bX^\top \bigg(\sum_{n=1}^N \left[ \mathbb{E}[\noisybx_n] \mathbb{E}[\noisybx_n]^\top + \diag\left(\mathbb{V}[\noisybx_n]\right) \right] \bigg)^{-1} \bigg(\sum_{n=1}^N \mathbb{E}\left[\noisybx_n \right] y_n\bigg) = \bH\by,
\end{equation}
where $\bH$ is often referred to as the hat matrix. If a single target value $y_n$ is replaced by $\hat{y}_n^{-n}$, the prediction made by the model trained on all data except example $n$, the hat matrix $\bH$ does not change because it does not depend on $\by$. If we construct a new target vector $\bz$ with this replacement in it, $\bz =[y_1, \dots, y_{n-1}, \hat{y}_n^{-n}, y_{n+1}, \dots, y_N]$, and use the simple fact that $\hat{y}_n^{-n}=\sum_{m=1}^N H_{nm} z_m$ (see \citet{allen74} for details) we obtain:
\begin{equation}
\hat{y}_n - \hat{y}_n^{-n} = \sum_{m=1}^N H_{nm} y_m - \sum_{m=1}^N H_{nm} z_m = H_{nn} y_n - H_{nn} \hat{y}_n^{-n}.
\end{equation}
Using this equality, we can compute $\hat{y}_n^{-n}$ without performing the minimization of the MCF quadratic loss -- with example $n$ left out -- over $\bw$:
\begin{equation}
\hat{y}_n^{-n} = \frac{H_{nn} y_n - \hat{y}_n}{H_{nn} - 1}.
\end{equation}
This allows us to express the leave-one-out error $LOO$ of the MCF quadratic loss model as:
\begin{equation}
LOO = \frac{1}{N} \sum_{n=1}^N \left(y_n - \hat{y}_n^{-n} \right)^2 = \frac{1}{N} \sum_{n=1}^N \left(\frac{y_n - \hat{y}_n}{1 - H_{nn}} \right)^2.
\end{equation}

\begin{table}[t]
\begin{center}
\resizebox{1.0\textwidth}{!}{
\begin{tabular}{|l|l|l|}\hline
\bf Distribution & $\boldsymbol{\mathbb{E}[\exp(-y_n w_d \noisyx_{nd})]_{p(\noisyx_{nd} | x_{nd})}}$ & $\boldsymbol{\partial \mathbb{E}[\exp(-y_n w_d \noisyx_{nd})]_{p(\noisyx_{nd} | x_{nd})} / \partial w_d}$\\\hline
Blankout noise & $q_d + (1 - q_d) \exp(-y_n w_d \frac{1}{1 - q_d} x_{nd})$ & $-\exp(-y_n w_d \frac{1}{1 - q_d} x_{nd}) y_n x_{nd}$\\
\rowcolor{gray} Gaussian noise & $\exp(-y_n w_d x_{nd} + \frac{1}{2}\sigma^2 y_n^2 w_{d}^2)$ & $\exp(-y_n w_d x_{nd} + \frac{1}{2}\sigma^2 y_n^2 w_{d}^2) (-y_n x_{nd} + \sigma^2 y_n^2 w_d)$\\
Laplace noise & $(1 - \lambda^2 y_n^2 w_d^2)^{-1} \exp(-y_n w_d x_{nd})$ & \pbox{10cm}{$-\exp(-y_n w_d x_{nd}) [(1-\lambda^2 y_n^2 w_d^2)^{-1} y_n x_{nd} +$ \\ $~~~~~~~~~~~~~~~~~~~~~~~~~~~~~2(1 - \lambda^2 y_n^2 w_d^2)^{-2} \lambda^2 y_n^2 w_d ]$}\\
\rowcolor{gray} Poisson noise & $\exp(x_{nd} (\exp(-y_n w_d) - 1))$ & $-\exp(x_{nd} (\exp(-y_n w_d) - 1) - y_n w_d) y_n x_{nd}$\\\hline
\end{tabular}
}
\caption{Moment-generating functions (MGFs) of various corrupting distributions, and the corresponding gradients of these functions with respect to the weights $w_d$. These quantities can be plugged into equations (\ref{eq:final_exp}) and (\ref{eq:final_log}) to obtain the expected value of the loss (or surrogate) under the corrupting distribution of the exponential and logistic loss functions, respectively.}\label{table:overview2}
\end{center}
\end{table}

\subsection{Exponential loss} 
\label{Exponential loss}
Whilst quadratic losses are of interest to regression problems, they seem less appropriate for use in classification. Exponential loss is a loss function that is commonly used for classification, \emph{e.g.}, in AdaBoost \citep{freund95}. Assuming a label variable $y \in \{0,1\}$, the expected value of the exponential loss under corruption model $p(\noisybx | \bx)$ is given by:
\begin{align}
\mathcal{L}(\mathcal{D}; \bw) &= \frac{1}{N} \sum_{n=1}^N \mathbb{E}\left[ \exp\left(-y_n \bw^\top\noisybx_n\right)\right]_{p(\noisybx_n | \bx_n)}\nonumber\\
&= \frac{1}{N} \sum_{n=1}^N \prod_{d=1}^D \mathbb{E}\left[\exp\left(-y_n w_d \noisyx_{nd}\right)\right]_{p(\noisyx_{nd} | x_{nd})},\label{eq:final_exp}
\end{align}
which we refer to as MCF exponential loss. Note that in the derivation, we used the assumption that the corruption is independent across features. The above equation can be recognized as a product of moment-generating functions $\mathbb{E}[\exp(t_{nd} \noisyx_{nd})]$ with $t_{nd} = -y_n w_d $. By definition, the moment-generating function (MGF) can be computed for all corrupting distributions that are member of the natural exponential family. An overview of the moment-generating functions for some corrupting distributions of interest is given in Table~\ref{table:overview2}. Noting that $f(x)$ is convex iff $\forall x: \frac{\partial^2 f(x)}{\partial x^2} \geq 0$ and noting that sums of convex functions are themselves convex, it is straightforward to verify that MCF exponential loss is convex for all corrupting distributions. We also note that for corrupting distributions that do not permit a moment-generating function but that have bounded support, it may be practical to upper bound (\ref{eq:final_exp}) using Hoeffding's lemma; we leave such investigations to future work.

The derivation of (\ref{eq:final_exp}) can readily be extended to a multi-class exponential loss with $K$ classes by replacing the weight vector $\bw$ by a $D \times K$ weight matrix $\mathbf{W}$, and by replacing the labels $y$ by label vectors $\by = \{1, -\frac{1}{K - 1} \}^K$ with $\sum_{k=1}^K y_k = 0$ \citep{zhu06}.

Unlike the minimization of MCF quadratic loss, the minimization of MCF exponential losses cannot be done in closed form and needs to be performed using gradient-descent techniques. Motivated by \citet{sha03}, we used an L-BFGS optimizer to minimize MCF exponential loss in this study.

\paragraph{Blankout corruption.} As an illustrative example, we work out MCF exponential loss with blankout corruption:
\begin{align}
\mathcal{L}(\mathcal{D}; \bw) = \frac{1}{N} \sum_{n=1}^N \prod_{d=1}^D \left(q_d + (1 - q_d) \exp(-y_n w_d x_{nd})\right).\label{eq:exp_blankout}
\end{align}
The gradient of this loss function is given by:
\begin{align}
\frac{\partial \mathcal{L}}{\partial w_d} = -\frac{1}{N} \sum_{n=1}^N (1 - q_d) \exp(-y_n w_d x_{nd}) y_n x_{nd} \prod_{d' \neq d} \left(q_{d'} + (1 - q_{d'}) \exp(-y_n w_{d'} x_{nd'}) \right).
\end{align}
Although the complexity of training with MCF remains $\mathcal{O}(N)$, evaluating this gradient requires more computation than evaluating the gradient of a standard exponential loss: in practice, the computation of the MCF exponential loss gradient is about twice as expensive when computation from the loss (\ref{eq:exp_blankout}) is reused efficiently.

\section{MCF for Logistic Loss}
\label{sec:upperlog}
\label{Logistic loss}

In the case of the logistic loss, the solution to the expected loss~(\ref{eq:regret}) cannot be computed in closed form. Instead, we derive one approximation and two practical upper bounds for the expected logistic loss, all of which can be used as surrogate loss functions in practice. In this section we follow the framework of~\cite{wagerArxiv13}, which is clearer than our previously used derivation~(\cite{vandermaaten13}) and leads to a nice interpretation of MCF as an adaptive regularizer. 

We will focus on binary classification in our derivations below, but our results can readily be extended to multi-class logistic loss. (In general, this is achieved by redefining the labels $y$ to be label vectors of the form $\by \in \{0, 1\}^K$ with $\sum_{k=1}^K y_k = 1$; and by defining the loss as the logarithm of the softmax probability of the correct prediction.)

\subsection{MCF Regularizer}
Assuming labels $y \in \{-1, +1\}$, we may define the logistic regression model as a generalized linear model, $p(y|x;\bw)=\frac{\exp(y\bw^\top\bx)}{\exp(-\bw^\top\bx)+\exp(\bw^\top\bx)}$. The negative log-likelihood, or the \emph{logistic loss}, of an instance pair $(\bx,y)$ under this model can be written as:
\begin{equation}
	L(\bx,y,\bw)=
	-y\bw^\top\bx+A(\bw^\top\bx) \textrm{ with: } A(\bw^\top\bx)=\log\left[\exp\left(-\bw^\top\bx\right)+\exp\left(\bw^\top\bx\right) \right].\label{eq:loglosswagner}
\end{equation}
Here, $A(\bw^\top \bx)$ is the log-partition function, which is independent of the label $y$. 
Noting that we assumed the corruption to be unbiased (\emph{i.e.\ } $\mathbb{E}[\noisybx]=\bx$), we can plug (\ref{eq:loglosswagner}) into the expected loss (\ref{eq:regret}) and obtain:
\begin{align}
	\mathcal{L}(\mathcal{D}; \bw)&=	  -\frac{1}{N} \sum_{n=1}^N \mathbb{E}\left[
	y_n\bw^\top\noisybx_n-A(\bw^\top\noisybx_n)
	\right]_{p(\noisybx_n | \bx_n)}\nonumber\\
	&=\frac{1}{N}\sum_{n=1}^N \left( \underbrace{L(\bx_n,y_n,\bw)}_{\textrm{log-loss}} +\underbrace{\mathbb{E}\left[A(\bw^\top\noisybx_n)\right]_{p(\noisybx_n | \bx_n)} - A(\bw^\top\bx_n)}_{\textrm{regularizer}}\right).\label{eq:wagnerreg}
\end{align}
\cite{wagerArxiv13} observe that the first part of (\ref{eq:wagnerreg}) is the regular logistic loss without MCF, and the 
 second part  can be interpreted as an adaptive regularizer that satisfies several appealing properties: 1. it only depends on $\bw$ and $\bx_n$ but not on the label $y$; 2. it is always non-negative---which follows directly from the fact that $A(\bw^\top\bx)$ is convex and an application of Jensen's inequality, \emph{i.e.\ } $\mathbb{E}[A(\bw^\top\noisybx)]\geq A(\mathbb{E}[\bw^\top\noisybx])=A(\bw^\top\bx)$; and 3. it has an intuitive interpretation, as it minimizes the difference between the expected prediction before and after corruption---in other words, the score that the classifier assigns to an example $\bx_n$ should be robust with respect to the corruption distribution.

Unfortunately, $\mathbb{E}[A(\bw^\top\noisybx)]$ in (\ref{eq:wagnerreg}) cannot be expressed in closed form. In the following three sections we therefore derive practical closed-form approximations and upper bounds that can be used to form surrogate loss functions for (\ref{eq:wagnerreg}).

\subsection{Quadratic Approximation}
\label{Quadratic Approximation}
A quadratic approximation to the expected logistic loss can be obtained by deriving a second-order Taylor expansion of $A(\bw^\top\noisybx)$ around $\bw^\top\bx$ and working out the expectation \citep{wagerArxiv13}. The resulting quadratic approximation to the expected logistic loss then becomes:
\begin{align}
	\mathcal{L}(\mathcal{D}; \bw) &\approx \frac{1}{N} \sum_{n=1}^N \left[-y_n\bw^\top\bx_n + A(\bw^\top \bx_n) + \frac{1}{2}\frac{\partial^2 A(\bw^\top \bx_n)}{\partial (\bw^\top \bx_n)^2}\mathbb{V}[\bw^\top \noisybx_n]_{p(\noisybx_n | \bx_n)}\right]\\
	&=\frac{1}{N} \sum_{n=1}^N \left[L(\bx_n,y_n,\bw) + \frac{1}{2}\frac{\partial^2 A(\bw^\top \bx_n)}{\partial (\bw^\top \bx_n)^2}\mathbb{V}[\bw^\top \noisybx_n]_{p(\noisybx_n | \bx_n)}\right].
	\label{eq:quadratic_approx}
\end{align}
In this equation, the first-order term has vanished because we assumed the corrupting distribution to be unbiased, $\mathbb{E}[\noisybx] = \bx$. The expression $\mathbb{V}[\bw^\top \noisybx_n]_{p(\noisybx_n | \bx_n)}$ denotes the variance of $\bw^\top \noisybx_n$ under the corrupting distribution, which takes the form:
\begin{equation}
\mathbb{V}[\bw^\top \noisybx_n]_{p(\noisybx_n | \bx_n)} = \bw^\top \left(\mathbb{E}[\noisybx_n \noisybx_n^\top]_{p(\noisybx_n | \bx_n)} - \bx_n\bx_n^\top \right)\bw.\label{eq:variance_quadratic}
\end{equation}
The second derivative of the log-partition function takes the form:
\begin{equation}
\frac{\partial^2 A(\bw^\top \bx_n)}{\partial (\bw \bx_n)^2} = \frac{4 \exp(-2 \bw^\top \bx_n)}{(1 + \exp(-2 \bw^\top \bx_n))^2} = 4 \sigma(2 \bw^\top \bx_n) (1 - \sigma(2 \bw^\top \bx_n))= \mathbb{V}[\hat{y}_n],
\end{equation}
where $\sigma(z)$ denotes the sigmoid function, $\sigma(z) = \frac{1}{1 + \exp(-z)}$, and $\mathbb{V}[\hat{y}_n]$ denotes the variance of the prediction $\hat{y}_n=\sign(\bw^\top\bx_n)$ under the logistic regression model\footnote{Please note that the mean $\mathbb{E}[\hat{y}] = \frac{\partial A(\bw^\top \bx)}{\partial (\bw \bx)} = 2 \sigma(2 \bw^\top \bx) - 1$.}. Hence, the quadratic approximation (\ref{eq:quadratic_approx}) can be interpreted as the sum of the standard logistic loss and a data-dependent, additive regularizer that is a product of: 1. the variance of the classifier score under the data corruption and 2. the variance of the prediction under the logistic regression model. Because the regularizer is the product of two variances, it is guaranteed to be non-negative. The additive regularizer vanishes when no corruption is applied on the data, because (\ref{eq:variance_quadratic}) is zero when the data is not corrupted.

\subsection{Jensen's upper bound}\label{Jensen's upper bound}

As an alternative to the quadratic approximation, we can obtain an effective upper bound on the expected logistic loss under the corrupting distribution using Jensen's inequality, which states that, because $\log(\cdot)$ is a concave function, $\mathbb{E}[\log(z)]\leq \log(\mathbb{E}[z])$. In particular, we obtain the following upper bound on the log-partition function:
\begin{equation}
\mathbb{E}\left[\log\left(\exp\left(-\bw^\top\noisybx\right)+\exp\left(\bw^\top\noisybx\right) \right)\right]\leq 
\log
\left(
\mathbb{E}\left[\exp\left(-\bw^\top\noisybx\right)\right]+
\mathbb{E}\left[\exp\left(\bw^\top\noisybx\right)\right]
\right),\label{eq:jensenA}
\end{equation}
where all expectations are under the corrupting distribution $p(\noisybx | \bx)$.

An alternative, somewhat simpler way to express the Jensen's upper bound on $\mathcal{L}(\mathcal{D}; \bw)$ can be achieved by redefining the labels $y \in \{0, 1\}$, and rephrasing (\ref{eq:wagnerreg}) as: 
\begin{equation}
\mathcal{L}(\mathcal{D}; \bw)= \frac{1}{N} \sum_{n=1}^N \mathbb{E}\left[\log\bigg(1 + \exp\left(-y_n \bw^\top \noisybx_n\right)\bigg)\right]_{p(\noisybx_n | \bx_n)},\label{eq:logistic}	
\end{equation}

Applying Jensen's inequality for $\log(\cdot)$ to (\ref{eq:logistic}), and subsequently following the same steps as in (\ref{eq:final_exp}), leads to:
\begin{equation}
\mathcal{L}(\mathcal{D}; \bw)\leq \frac{1}{N} \sum_{n=1}^N \log\bigg(1 + \prod_{d=1}^D \mathbb{E}\left[\exp\left(-y_n w_d \noisyx_{nd}\right)\right]_{p(\noisyx_{nd} | x_{nd})}\bigg).\label{eq:final_log}
\end{equation}
We again recognize a product of MGFs, which can be computed in closed-form for corrupting distributions in the natural exponential family (see Table~\ref{table:overview2}). The upper bound on the expected logistic loss (\ref{eq:final_log}) can be rewritten as a log-sum-exp or log-integral-exp function (depending on whether the corruption is discrete or continuous) of an affine function of the parameters $\mathbf{w}$, as a result of which it is guaranteed to be convex \citep[\S 3.1.5]{boyd04}. We note here that the evaluation of (\ref{eq:final_log}) in a numerically stable way is non-trivial when blankout corruption is used; for details on numerical stability, we refer to Appendix~\ref{Numerical stability}.

\paragraph{Poisson corruption.} As an illustrative example, we show the upper bound of the logistic loss (\ref{eq:final_log}), where inputs are corrupted with the Poisson distribution:
\begin{equation}
\mathcal{L}(\mathcal{D}; \bw)\leq \frac{1}{N} \sum_{n=1}^N \log\left(1+\exp\bigg({\sum_{d=1}^Dx_{nd} (\exp(-y_n w_d) - 1)\bigg)}\right).\label{eq:logistic_poisson}
\end{equation}
As with all MCF losses that use Poisson corruption, this regularized loss does not have any additional hyper-parameters. Further, the sum over all features can still be computed efficiently for sparse data by summing only over non-zero entries in $\bx_n$. As for MCF exponential loss, we perform the minimization of this surrogate loss with an L-BFGS optimizer. The  gradient of (\ref{eq:logistic_poisson}) is given by:
\begin{equation}
\frac{\partial \mathcal{L}}{\partial w_d} = - \frac{1}{N} \sum_{n=1}^N \frac{\exp\left(\sum_{d' \neq d} x_{nd'} (\exp(-y_n w_{d'}) - 1)\right)}{1 + \exp\left(\sum_{d=1}^D x_{nd} (\exp(-y_n w_d) - 1)\right)} \exp(y_n w_d) y_n x_{nd}. 
\end{equation}
It is fair to say that in practice, the computation of the gradient of this MCF logistic loss takes approximately $10\times$ longer than the computation of a regular logistic loss gradient (due to extra exponentiations and multiplications in the gradient computation).

\subsection{Variational bound}\label{Variational bound}
\begin{figure}
\centerline{\includegraphics[width=0.7\textwidth]{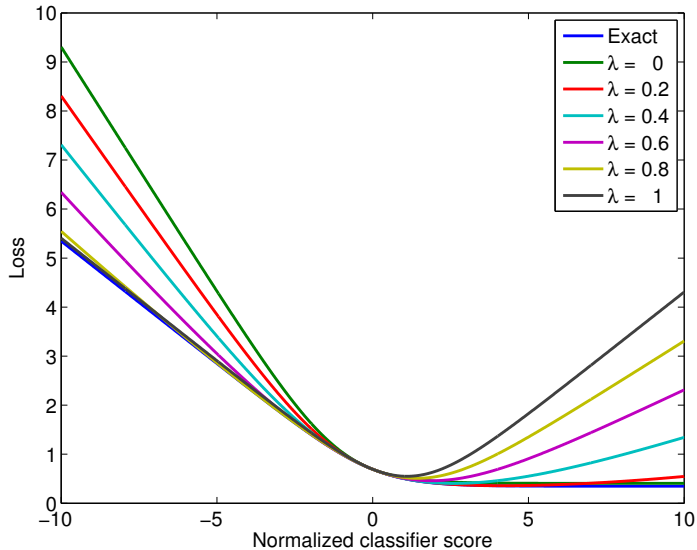}}
\caption{Variational upper bounds on the MCF-blankout logistic loss with $q=0.5$ for various values of $\lambda$. Please note that the Jensen's upper bound of Section~\ref{Jensen's upper bound} corresponds to the case $\lambda=0$.}
\label{fig:variational_bounds}
\end{figure}

Indeed, Jensen's inequality is not the only bound that may be used to derive an upper bound on MCF logistic loss. In particular, we also consider variational bounds of the form:
\begin{align}
\mathcal{L}(\mathcal{D}; \bw) &= \frac{1}{N} \sum_{n=1}^N \mathbb{E}\left[\log(1 + \exp(-y_n \bw^\top \noisybx_n))\right]_{p(\noisybx_n | \bx_n)}\nonumber\\
&=\min_{\lambda\in{[0,1]}} \frac{1}{N} \sum_{n=1}^N \left( \mathbb{E}\left[-\lambda y_n \bw^\top \noisybx_n\right]_{p(\noisybx_n | \bx_n)} + \mathbb{E}\left[\log(\exp(\lambda y_n \bw^\top \noisybx_n) (1 + \exp(-y_n \bw^\top \noisybx_n)))\right]\right)\nonumber\\
&\leq \min_{\lambda\in{[0,1]}}\frac{1}{N} \sum_{n=1}^N \left(\log\left(\mathbb{E}\left[\exp(\lambda y_n \bw^\top \noisybx_n)\right] + \mathbb{E}\left[\exp((\lambda - 1) y_n \bw^\top \noisybx_n)\right]\right) - \lambda y_n \bw^\top \bx_n\right),\label{eq:tighter_bound}
\end{align}
where we use the assumption that the corrupting distribution is unbiased, $\mathbb{E}[\noisybx] = \bx$. The expression above contains two moment-generating functions (with $t = \lambda y_n \bw$ and $t = (\lambda-1) y_n \bw$, respectively), which can be evaluated as before. Minimizing (\ref{eq:tighter_bound}) over the hyperparameter $\lambda \in [0, 1]$, using a grid or gradient search, leads to a potentially tighter bound than that in (\ref{eq:final_log}). Note that the special case $\lambda = 0$ equals  (\ref{eq:final_log}), as a result of which the alternative bound is always at least as tight. The bounds are illustrated for various values of $\lambda$ in Figure~\ref{fig:variational_bounds}.

\subsection{Graphical model interpretation} 
\begin{figure}
\centerline{\includegraphics[width=0.45\textwidth]{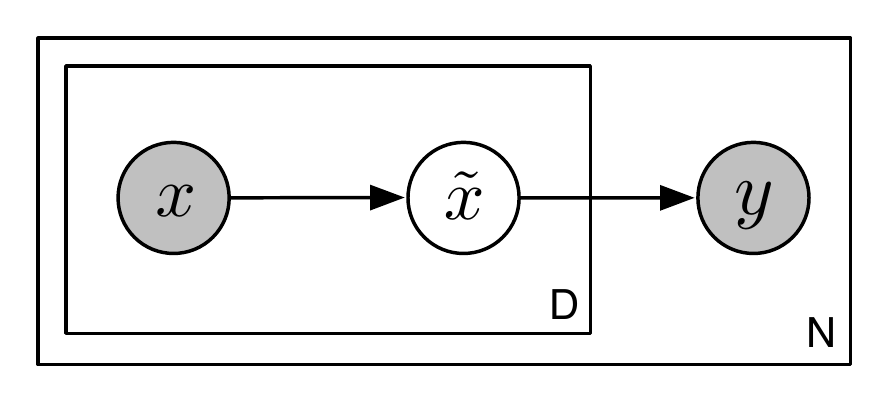}}
\caption{Graphical model interpretation of marginalized corrupted features (MCF). Shaded circles represent observed variables; non-shaded circles represent latent variables; and plates denote independent repetitions. The arrow from $x_{nd}$ to $\noisyx_{nd}$ represents the corrupting distribution $p(\noisyx_{nd} | x_{nd})$ and the arrow from $\noisybx_n$ to $y_n$ represents the logistic regression model $p(y_n|\noisybx_n)$.}
\label{fig:bayes_net}
\end{figure}

Finally, minimizing MCF logistic loss can also be interpreted as training the parameters of the simple Bayesian network in Figure~\ref{fig:bayes_net} via maximum likelihood. In this network, $p(\noisyx_d | x_d)$ denotes a corrupting distribution, and $p(y | \noisybx)$ comprises a simple linear logistic regressor: $p(y | \noisybx)=1 / (1 + \exp(-y\bw^\top\noisybx))$. Marginalizing out the corrupted data $\noisybx$, we obtain:
\begin{align}
p(y_n | \bx_n) = \mathbb{E}\left[\frac{1}{1 + \exp(-y_n \bw^\top\noisybx_n)}\right]_{p(\noisybx_n | \bx_n)}\nonumber\\
\geq \frac{1}{1 + \mathbb{E}\left[\exp(-y_n \bw^\top\noisybx_n)\right]_{p(\noisybx_n | \bx_n)}},\label{eq:bayes_net}
\end{align}
where we have used the fact that $\frac{1}{1+z}$ is convex on $[0, \infty)$ and that $\forall z: 1 + \exp(z) > 1$ to lower bound $p(y | \bx)$ using Jensen's inequality. (Indeed, we could also have applied the quadratic approximation of~\ref{Quadratic Approximation} or the variational approximation of~\ref{Variational bound} here.)

Because the corrupting distribution $p(\noisybx | \bx)$ factorizes over dimensions, the log-likelihood $\ell$ of data $\mathcal{D}$ can thus be lower bounded by:
\begin{align}
\ell(\mathcal{D}; \bw) \geq -\sum_{n=1}^N \log\left(1 + \prod_{d=1}^D \mathbb{E}\left[\exp(-y_n w_d \noisyx_{nd})\right]_{p(\noisyx_{nd} | x_{nd})}\right).
\end{align}
It is straightforward to verify that maximizing this lower bound on the log-likelihood is identical to minimizing (\ref{eq:final_log}). Further, it it possible to develop a similar graphical model for MCF-quadratic loss by setting $p(y_n | \noisybx_n) = \mathcal{N}(y_n | \bw^\top \noisybx, \sigma^2)$.

The graphical model interpretation of MCF also suggests an alternative way to make predictions with models trained using MCF, \emph{viz.} by evaluating (\ref{eq:bayes_net}). In practice, this amounts to generating infinitely many corrupted copies of the test data using $p(\noisybx | \bx)$
and \emph{averaging over the predictions for all corrupted test points}. While corrupting test data during prediction may sound counterintuitive at first, it does provide an elegant way to remove the restriction for MCF corruption distributions to be unbiased (especially in the case of regression). In classification, corrupting test data provides a natural way to measure the uncertainty of a prediction, even when the original predictor was non-probabilistic (much like conformal prediction; \citet{shafer2008tutorial}).

\section{Experiments}
\label{Experiments}


We evaluate MCF predictors on four tasks: 1. document classification based on word-count features using MCF with blankout and Poisson corruption; 2. image classification based on bag-of-visual-word features using MCF with blankout and Poisson corruption; 3. splice-junction recognition in DNA sequences using blankout and multinomial corruption; and 4. classification of objects in the ``nightmare at test time'' scenario using MCF with blankout corruption. The four sets of experiments are described separately below. Code to reproduce the results of our experiments is available online\footnote{A Matlab implementation is available at \url{http://bit.ly/11bn2GG}.}. 

\begin{figure*}[h!]
\centering
\includegraphics[width=.9\textwidth]{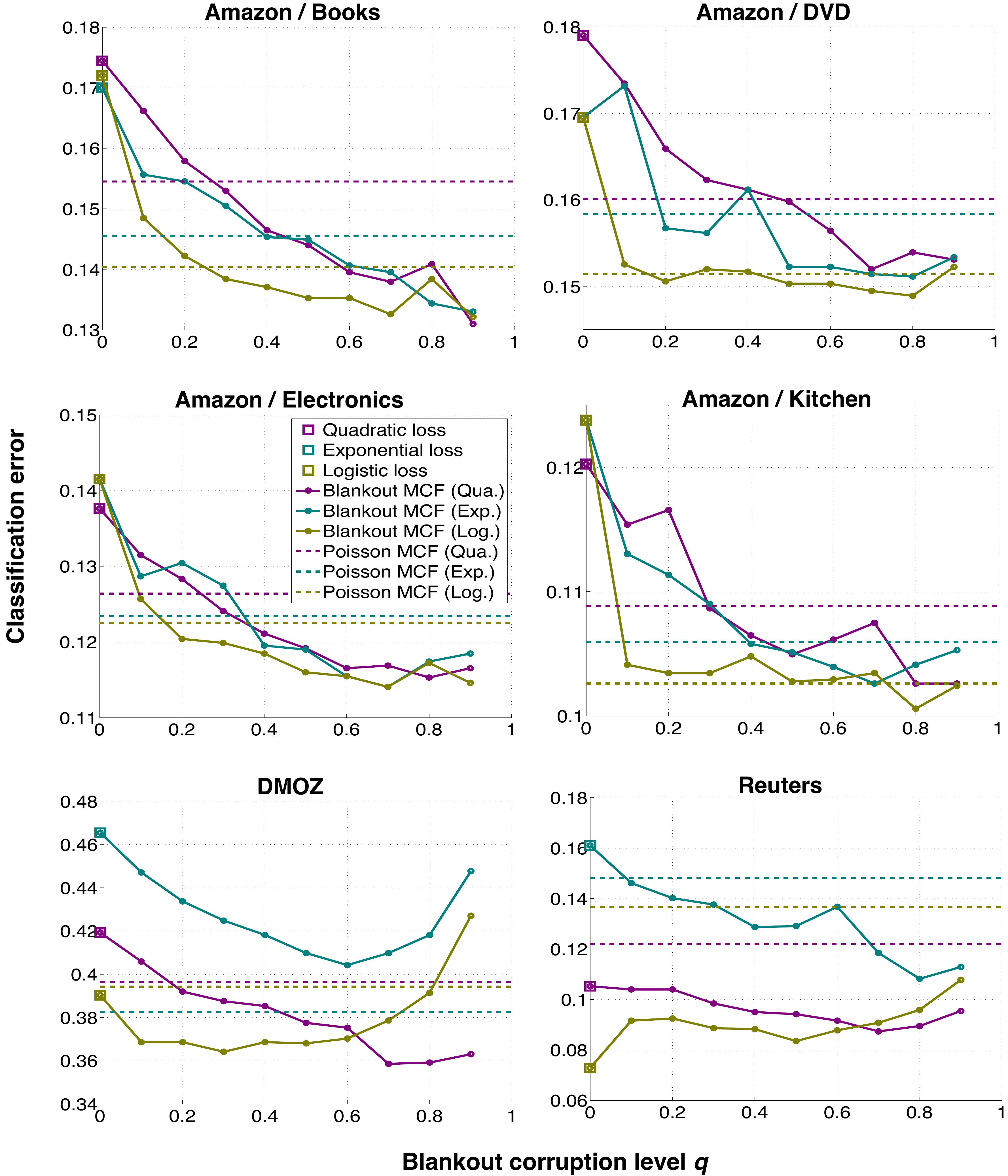}
\caption{Classification errors of MCF predictors using blankout and Poisson corruption -- as a function of the blankout corruption level $q$ -- on the Amazon data set for $l_2$-regularized quadratic, exponential, and quadratic loss functions. Classification errors are represented on the $y$-axis, whereas the blankout corruption level $q$ is represented on the $x$-axis. The case of MCF with blankout corruption and $q = 0$ corresponds to a standard $l_2$-regularized classifier. Figure best viewed in color.}
\label{fig:amazon}
\hspace{.015\textwidth}
\end{figure*}

\subsection{Document classification}
\label{Document classification}

We first test MCF predictors with blankout and Poisson corruption on document classification tasks. Specifically, we focus on three data sets: the Dmoz data set, the Reuters data set, and the Amazon review benchmark set~\citep{blitzer2007biographies}. 

\textbf{Data sets.} The Dmoz open directory (\url{http://www.dmoz.org}) contains a large collection of webpages arranged into a tree hierarchy. We use a subset consisting of $N = 8,980$ webpages from the $K = 16$ categories in the top level of the hierarchy. Each webpage is represented by a bag-of-words representation with $D = 16,498$ words. The Reuters document classification data set consists of $N = 8,293$ news articles that appeared on the Reuters newswire in 1987 belonging to $K = 65$ topics (documents corresponding to multiple topics were removed from the data). The bag-of-words representation contains $D = 18,933$ words for each document. The four Amazon data sets consist of approximately $N = 6,000$ reviews of four types of products: books, DVDs, electronics, and kitchen appliances. Each review is represented by a bag-of-words representation of the $D = 20,000$ most common words. On the Dmoz and Reuters data sets, the task is to classify the documents into one of the predefined categories. On the Amazon data set, the task is to decide whether a review is positive or negative. 

\textbf{Setup.} On the Dmoz and Reuters data sets, we use fixed $75\% / 25\%$ training/test splits. On the Amazon data set, we follow the experimental setup of~\citet{blitzer2007biographies} by using a predefined division of the data into approximately $2,000$ training examples and about $4,000$ test examples (the exact numbers vary slightly between tasks). We perform experiments with linear classifiers that are trained using $l_2$-regularized quadratic, exponential, and logistic loss functions. In all experiments, the amount of $l_2$-regularization is determined via cross-validation. The minimization of the (expected) exponential and logistic losses is performed by running Mark Schmidt's \url{http://www.di.ens.fr/~mschmidt/Software/minFunc.html}{\texttt{minFunc}}-implementation of L-BFGS until convergence or until a predefined maximum number of iterations is reached. In our experiments with MCF logistic loss, we used the Jensen's upper bound of Eqn.~(\ref{eq:final_log}).

All our predictors included a bias term that is neither regularized nor corrupted. In our experiments with MCF using blankout corruption, we use the same noise level for each feature, \emph{i.e.} we assume that $\forall d: q_d  =  q$. On all data sets, we first investigate the performance of MCF as a function of the corruption level $q$ (but we still cross-validate over the $l_2$-regularizer). In a second set of experiments, we cross-validate over the blankout corruption parameter $q$ and study to what extent the performance (improvements) of MCF depend on the amount of available training data. (MCF with Poisson corruption has no additional hyperparameters, as a result of which it requires no extra cross-validations.)

\begin{figure*}[t]
\centering
\includegraphics[width=.85\textwidth]{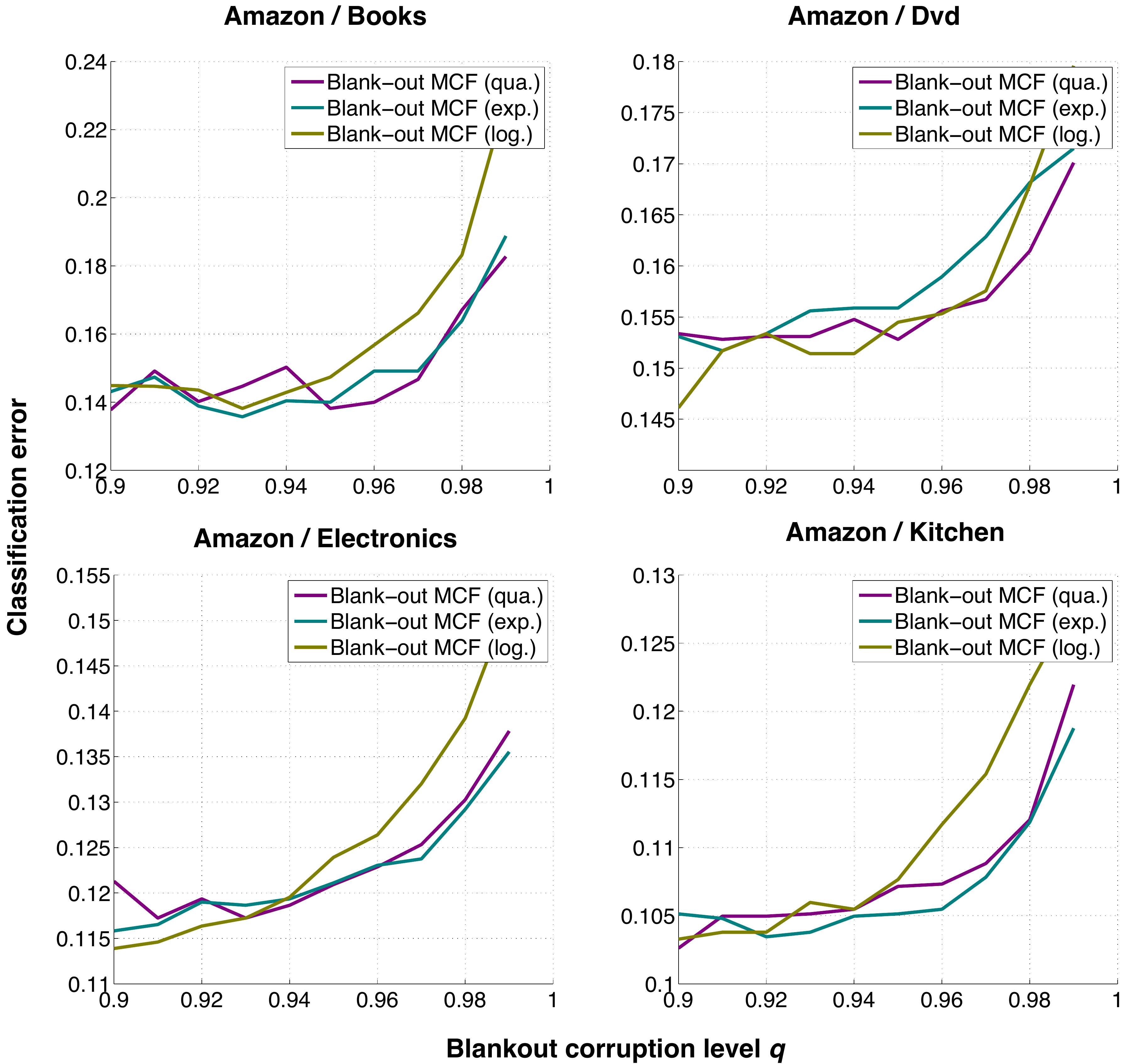}
\caption{Classification errors of MCF predictors using blankout corruption -- as a function of the blankout corruption level $q$ in the range of $[0.9, 1]$ on the Amazon review dataset. }
\label{fig:amazon-closeup}
\end{figure*}

\begin{figure}[htp]
\centerline{\includegraphics[width=\textwidth]{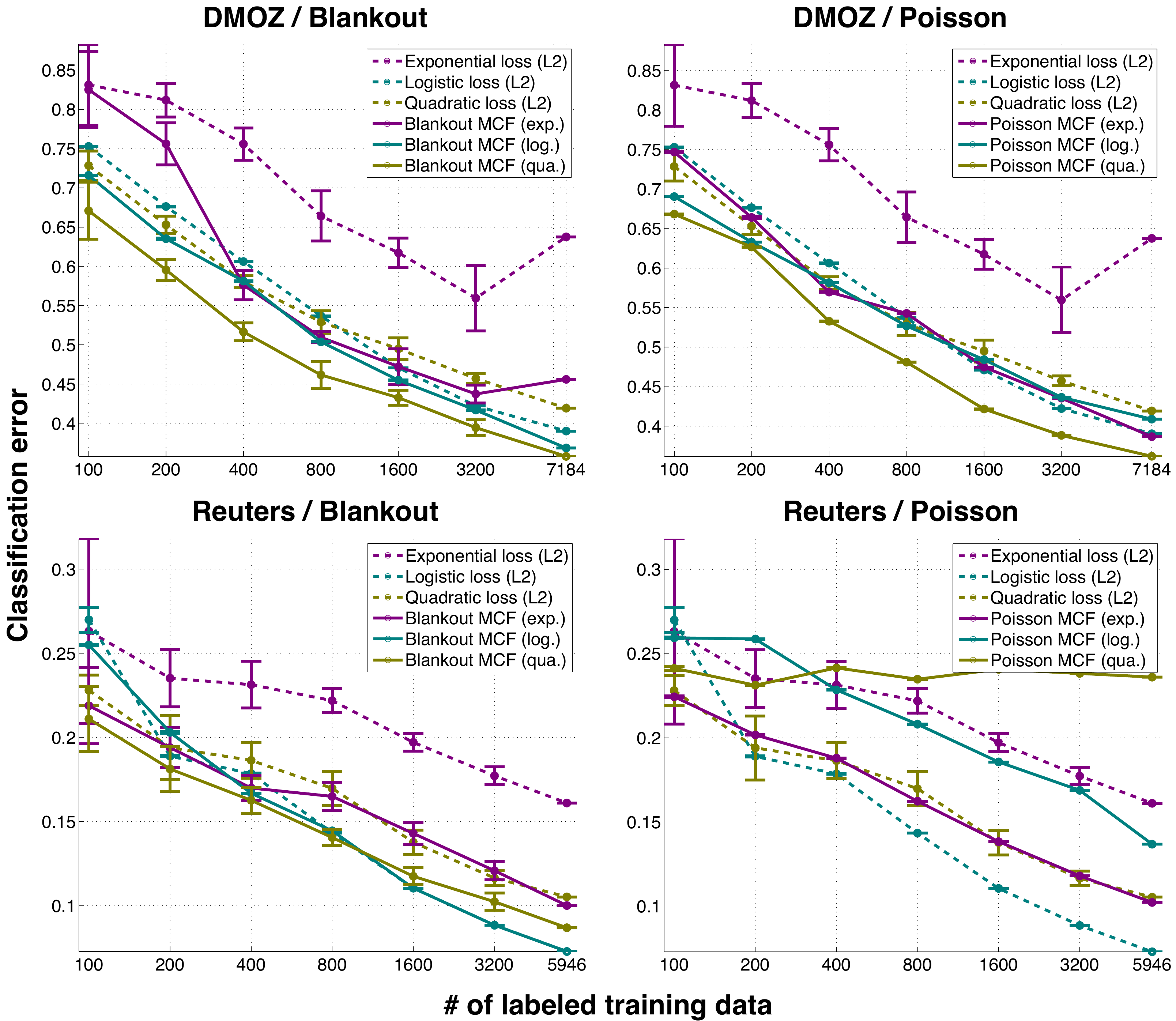}}
\caption{The performance of standard and MCF classifiers with blankout and Poisson corruption models as a function of training set size on the Dmoz and Reuters data sets. Both the standard and MCF predictors employ $l_2$-regularization. Figure best viewed in color.}
\label{fig:dmoz_reuters}
\hspace{.015\textwidth}
\end{figure}

\paragraph{Results.} Figure~\ref{fig:amazon} shows the test error of our MCF predictors on all data sets as a function of the blankout corruption level $q$. Herein, corruption level $q  =  0$ corresponds to the baseline predictors, \textit{i.e.} to predictors that do not employ MCF at all (but they do employ $l_2$-regularization). The results show: 1. that MCF consistently improves over standard predictors for both blankout corruption (for all corruption levels $q$) and Poisson corruption on five out of six tasks; 2. that MCF with Poisson corruption leads to substantial performance improvements over standard classifiers whilst introducing no additional hyperparameters; and 3. that the best performance tends to be achieved by MCF with blankout corruption with relatively high corruption levels are used. In particular, the best corruption level $q$ for the Amazon dataset is in the order of $0.8$ or $0.9$. Our explanation is that the ratio of relevant features (in this case, sentiment words) is fairly low in these data, and  higher dropout increases the robustness of the classifiers against irrelevant features. However, further increasing the corruption level will lead to over-regularization. As shown in Figure~\ref{fig:amazon-closeup}, the test error of MCF predictors increases drastically as the blankout corruption level exceeds $0.9$.  

The best-performing MCF classifiers reduce the test errors by up to $22\%$ on the Amazon data if $q$ is properly set. In many of the experiments with MCF-trained losses (in particular, when blankout corruption is used), we observe that the optimal level of $l_2$-regularization is $0$. This showcases the regularizing effect of MCF, which can render additional regularization superfluous.

Figure~\ref{fig:dmoz_reuters} presents the results of a second set of experiments on the Dmoz and Reuters data sets, in which we study how the performance of MCF depends on the amount of training data. For each training set size, we repeat the experiment five times with randomly sub-sampled training sets; the figure reports the mean test errors and the corresponding standard deviations. The results show that classifiers trained with MCF (solid curves) significantly outperform their counterparts without MCF (dashed curves). The performance improvement is consistent irrespective of the training set size, {\it viz.} up to $25\%$ on the Dmoz data set.

\begin{figure}
\centerline{\includegraphics[width=0.49\textwidth]{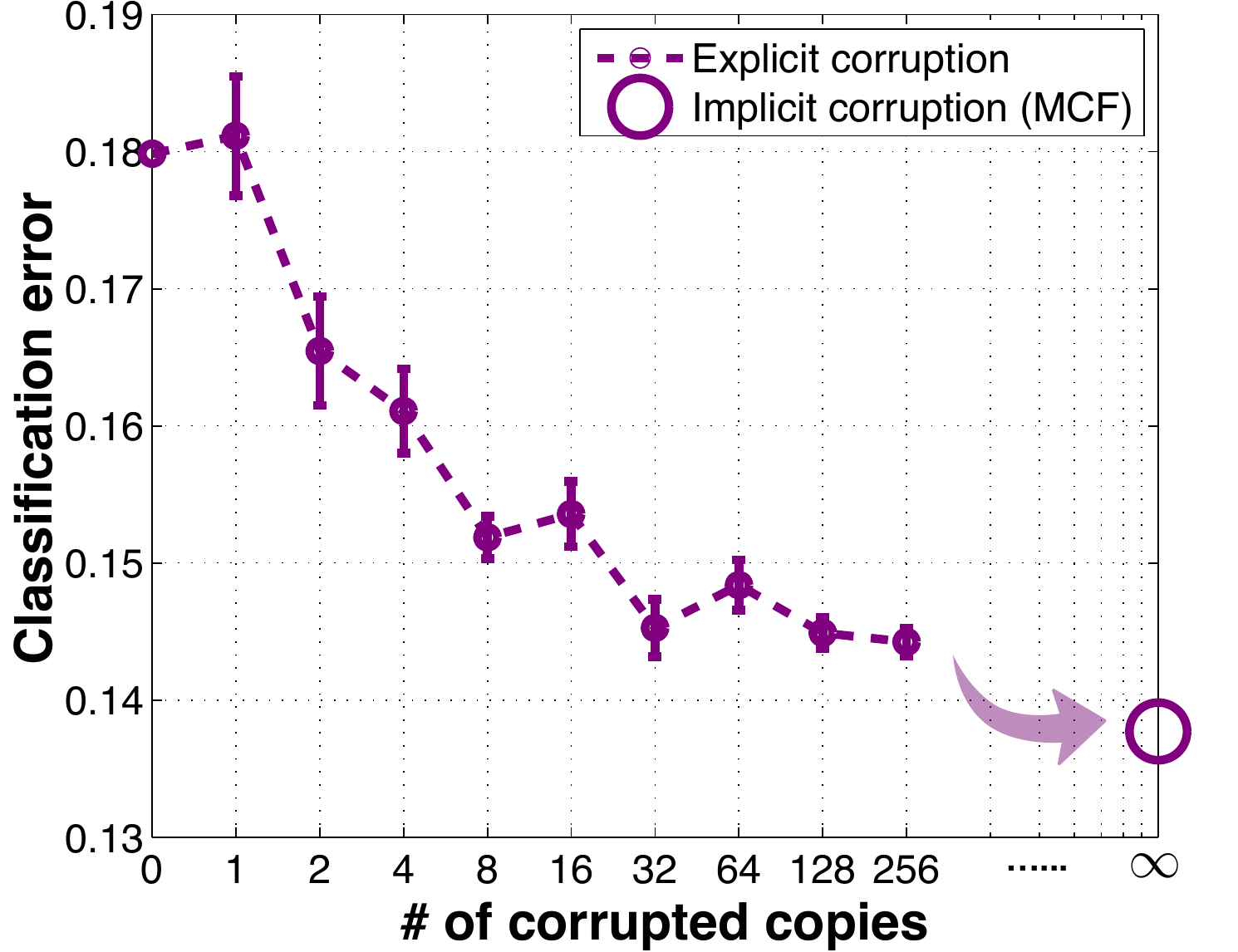}
\includegraphics[width=0.49\textwidth]{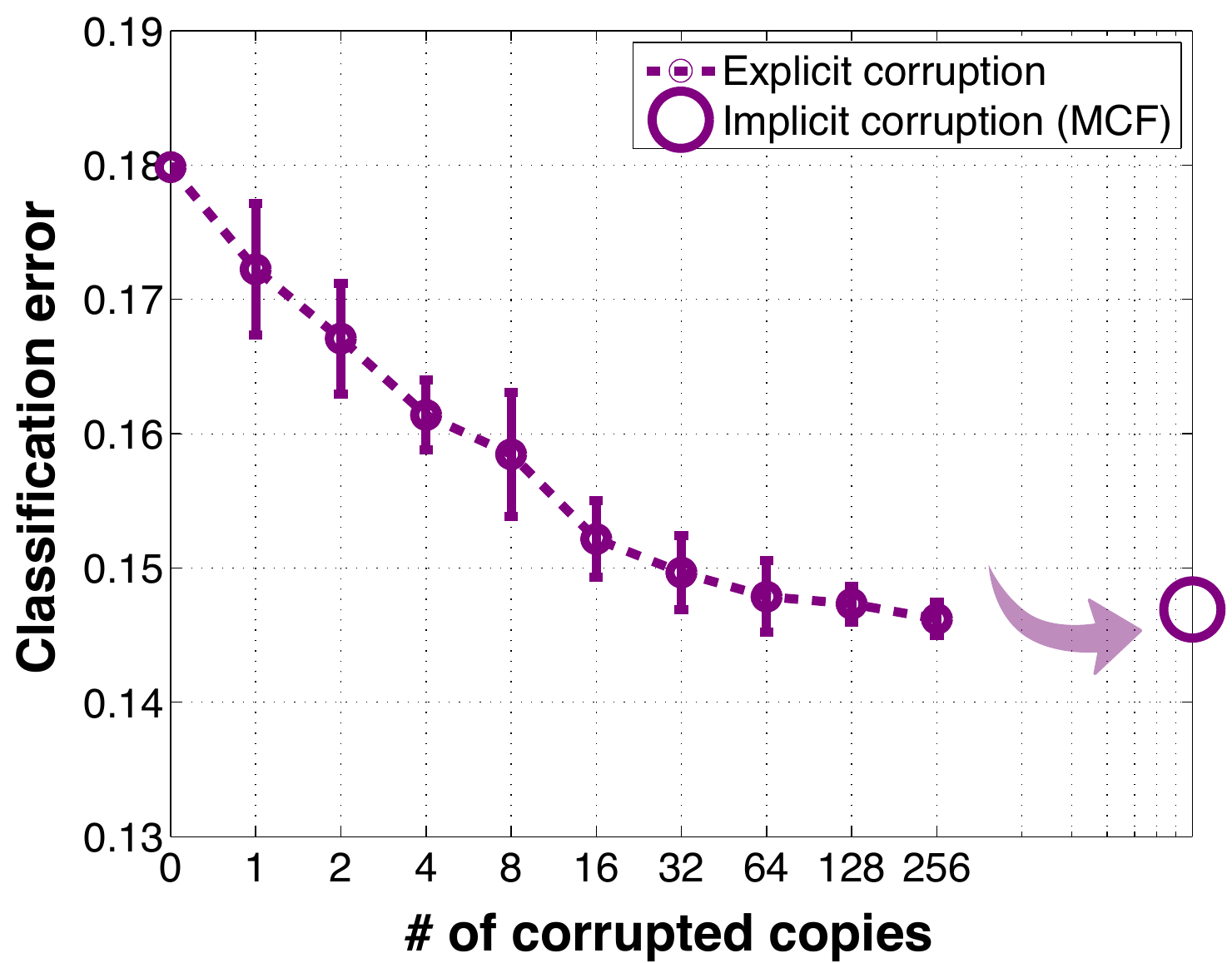}}
\caption{Comparison between MCF and explicitly adding corrupted examples to the training set (for quadratic loss) on the Amazon (books) data using blankout (\emph{left}) and Poisson (\emph{right} corruption. Training with MCF is equivalent to using infinitely many corrupted copies of the training data.
}
\label{fig:corrupt}
\vspace{-0.25cm}
\end{figure}

\textbf{Explicit vs. implicit feature corruption.} Figure~\ref{fig:corrupt} shows the classification error on the Amazon Books data set when a classifier without MCF is trained on the data set with additional \emph{explicitly} corrupted samples, as formulated in equation~(\ref{eq:explicitM}). To generate the results in the left graph of Figure~\ref{fig:corrupt}, we use the blankout corruption model with $q$ set by cross-validation for each training set size. To generate the right graph of Figure~\ref{fig:corrupt}, we used the Poisson corruption model (which has no hyperparameters). In both experiments, we trained the classifiers using quadratic loss and $l_2$-regularization (with cross-validation over the regularization parameter for each data size).

The graphs in the figure show a clear trend that the error \emph{decreases} when the training set contains more corrupted versions of the original training data, \emph{i.e.} with higher $M$ in equation~(\ref{eq:explicitM}). The graphs also illustrate that the best performance is obtained as $M$ approaches infinity, which is equivalent to MCF with blankout or Poisson corruption (big markers in the bottom right). This nicely illustrates the potential of MCF to improve generalization by training on corrupted examples without increasing the computational complexity of learning.

\begin{figure}[t]
\centerline{\includegraphics[width=0.85\textwidth]{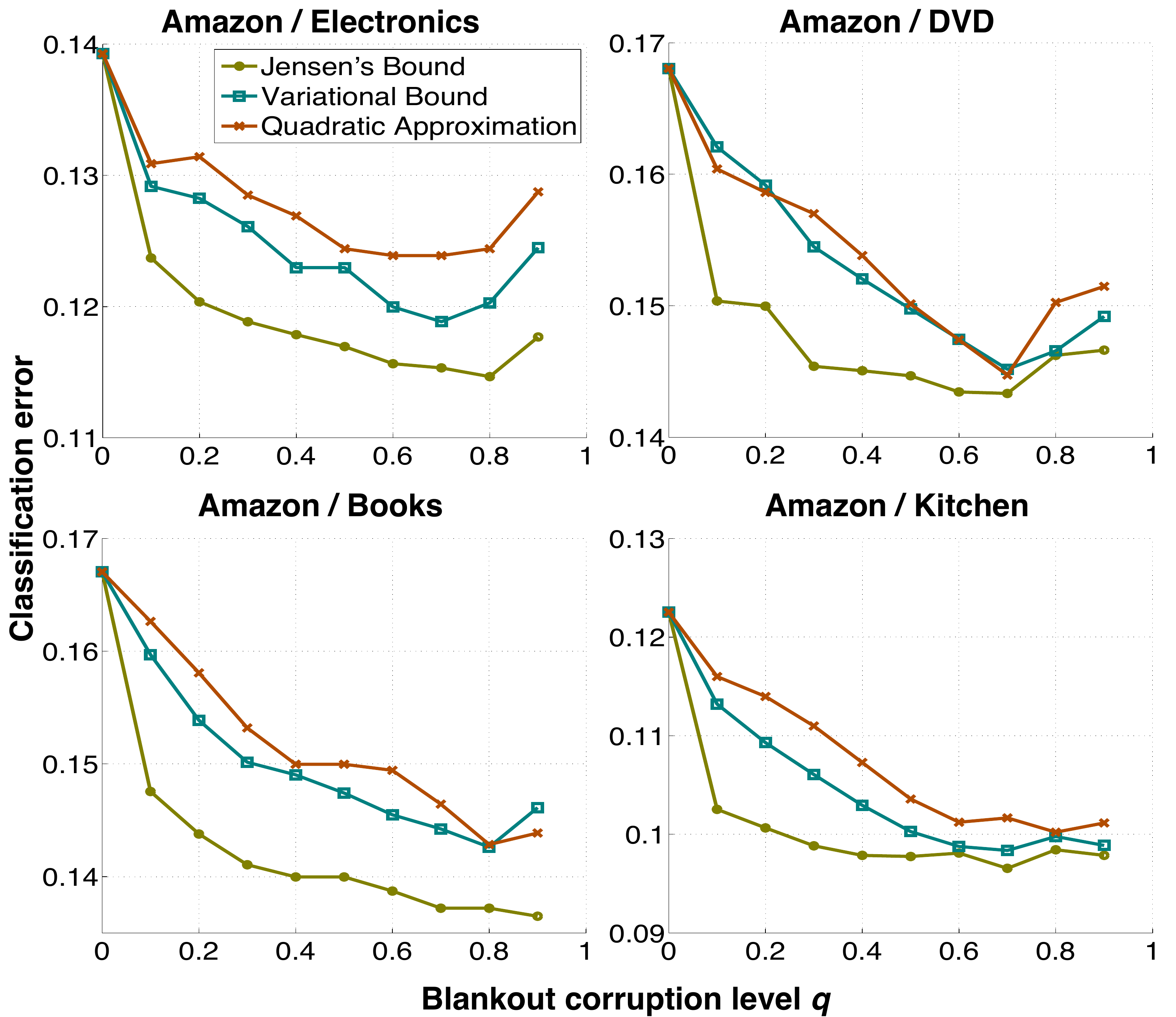}}
\caption{Comparison between the three approaches (quadratic approximation, Jensen's upper bound, and variational upper bound) to MCF logistic loss with blankout as a function of the blankout rate $q$. The value $q\!=\!0$ corresponds to classifiers trained using standard logistic loss. Figure best viewed in color.
}
\label{fig:log_approx}
\vspace{-0.25cm}
\end{figure}

\textbf{Comparing approximations and bounds for MCF logistic loss.} We also performed experiments to evaluate the effectiveness of the three approaches to deal with MCF logistic loss, \emph{viz.} the quadratic approximation of Eqn.~(\ref{eq:quadratic_approx}), the Jensen's upper bound of Eqn.~(\ref{eq:final_log}), and the variational upper bound of Eqn.~(\ref{eq:tighter_bound}). Specifically, we compared the performance of the three approaches when using blankout corruption with different values of $q$ on the four Amazon data sets. As before, all classifiers used additional $l_2$-regularization; the regularization parameters were set by cross-validating on a held-out validation set.

The results of these experiments are shown in Figure~\ref{fig:log_approx} (please note that, again, the setting $q\!=\!0$ corresponds to training classifiers using standard logistic loss). The results presented in the figure show that classifiers trained using logistic loss with MCF-blankout outperform classifiers that are trained using standard logistic loss, irrespective of whether the quadratic approximation or one of the two upper bounds is used. Having said that, the results also show that the two upper bounds on MCF logistic loss outperform the quadratic approximation in most experiments. In addition, the results show that the use of Jensen's upper bound tends to lead to better results than the use of the variational upper bound to the logistic loss, despite the fact that the variational bound is always at least as tight as Jensen's bound. We surmise this results is due to the fact that the variational bound of Eqn.~(\ref{eq:tighter_bound}) is not jointly convex in $\bw$ and $\lambda$, whereas the Jensen's bound of Eqn.~(\ref{eq:final_log}) is convex in $\bw$. As a result, we may get stuck in poorer local minima of the loss when using the variational upper bound.



\subsection{Image classification}
\label{Image classification}
We perform image-classification experiments with MCF on the CIFAR-10 data set \citep{krizhevsky09}, which is a subset of the 80 million tiny images \citep{torralba08}. The data set contains RGB images with 10 classes of size $32 \times 32$, and consists of a fixed training set of $50,000$ images and a fixed test set of $10,000$ images. 

\paragraph{Setup.} We follow the experimental setup of \citet{coates11}: we whiten the training images and extract a set of $7 \times 7$ image patches on which we apply $k$-means clustering (with $k = 2048$) to construct a codebook of prototypical image patches. Next, we slide a $7 \times 7$ pixel window over each image and identify the nearest prototype in the codebook for each window location. We construct an image descriptor by subdividing the image into four equally sized quadrants and counting the number of times each prototype occurs in each quadrant, which leads to a descriptor of dimensionality $D = 4 \times 2048$. This way of extracting the image features is referred to by \citet{coates11} as $k$-means with hard assignment, average pooling, patch size $7 \times 7$, and stride $1$. Because all images have the same size, we do not need to normalize the descriptors. We train MCF predictors with blankout and Poisson corruption on the full set of training images, cross-validating over a range of $l_2$-regularization parameters. The generalization error is evaluated on the test set. For our experiments with logistic loss, we used the Jensen's upper bound of Eqn.~(\ref{eq:final_log}).

\paragraph{Results.} The results are reported in Table~\ref{table:cifar_results}. The baseline classifiers (without MCF) are comparable to the $68.8\%$ accuracy reported by \citet{coates11} with exactly the same experimental setup (except for exponential loss). The results illustrate the potential of MCF classifiers to improve the prediction performance on bag-of-visual-words features, in particular, when using quadratic or logistic loss in combination with Poisson corruption. 

Although our focus in this section is to merely illustrate the potential of MCF on image classification tasks, it is worth noting that the best results in Table~\ref{table:cifar_results} match those of a highly non-linear mean-covariance RBMs trained on the same data \citep{ranzato10}, despite our use of very simple visual features and of linear classifiers. 

\begin{table}[t]
\begin{center}
\begin{tabular}{|l|r|r|r|}\hline
 & \bf Quadratic & \bf Exponential & \bf Logistic\\\hline
\bf No MCF & 32.6\% & 39.7\% & 32.5\%\\
\rowcolor{gray} \bf Poisson MCF & 29.1\% & 39.5\% & 30.0\%\\
\bf Blankout MCF & 32.3\% & 37.9\% & 29.4\%\\\hline
\end{tabular}
\caption{Classification errors obtained on the CIFAR-10 data set with MCF classifiers trained on simple spatial-pyramid bag-of-visual-words features.\vspace{-.5cm}}\label{table:cifar_results}
\end{center}
\end{table}


\subsection{Analysis of DNA}
We also performed experiments with MCF in which the aim is to automatically recognize splice-junctions in DNA: the locations that indicate the boundaries between regions of DNA that are used in the production of mRNA (so-called introns) and regions of DNA that are removed in the production of mRNA (exons). The two types of boundaries are called \emph{acceptors} (exon-to-intron boundaries) and \emph{donors} (intron-to-exon boundaries). The mRNA that is created using the information in the DNA is subsequently translated into proteins, which form the basis of all chemical processes inside the cell. Therefore, automatic recognition of splice-junctions is essential in the analysis of cell processes.

\paragraph{Setup.} We perform experiments on the data set constructed by \citet{noordewier91}. The data set comprises $3190$ examples of three classes: it contains $767$ examples of acceptors, $768$ examples of donors, and $1655$ negative examples. Each example is represented by a DNA subsequence of $120$ nucleotides: $60$ on each side of the boundary for positive examples (acceptors and donors). We represent the nucleotides using a $1$-of-$K$ representation, where $K=4$ is the number of distinct nucleotides (A, G, T, and C). This leads to binary feature vectors of dimensionality $D=480$.

We measure the performance of our classifiers using $10$-fold cross-validation. We repeat each experiment fifty times and average the performances across the fifty runs. In this experiment, we focus on MCF with logistic loss because in preliminary experiments, models trained with logistic loss produced substantially better results than models that used other loss functions. As in the previous experiments, we use the Jensen's upper bound on MCF-logistic loss of Eqn.~(\ref{eq:final_log}) as surrogate loss function. Because of the binary nature of our data, we focus on using blankout corruption in MCF. In addition, we explore a simple \emph{multinomial} corruption model that aims to model the sequencing errors that are made by DNA sequencing machines (\emph{e.g.}, \cite{le13,zagordi10}). The multinomial corruption model assumes that a base pair $x \in \{1, \dots, K\}$ is an erroneous read with probability $q$, and assumes a uniform distribution over all alternative base pairs in case of an erroneous read:
\begin{equation}
p(\noisyx = x) = 1-q \mbox{ and } \forall k \neq x: p(\noisyx = k) = \frac{q}{K-1}.
\end{equation}
We note here that this corrupting distribution violates the assumption that $\mathbb{E}[\noisybx] = \bx$. However, this is not a problem since we are focusing on a classification task (the unbiasedness assumption is only relevant to regression tasks). 

\begin{figure}[t]
\centerline{\includegraphics[width=.48\textwidth]{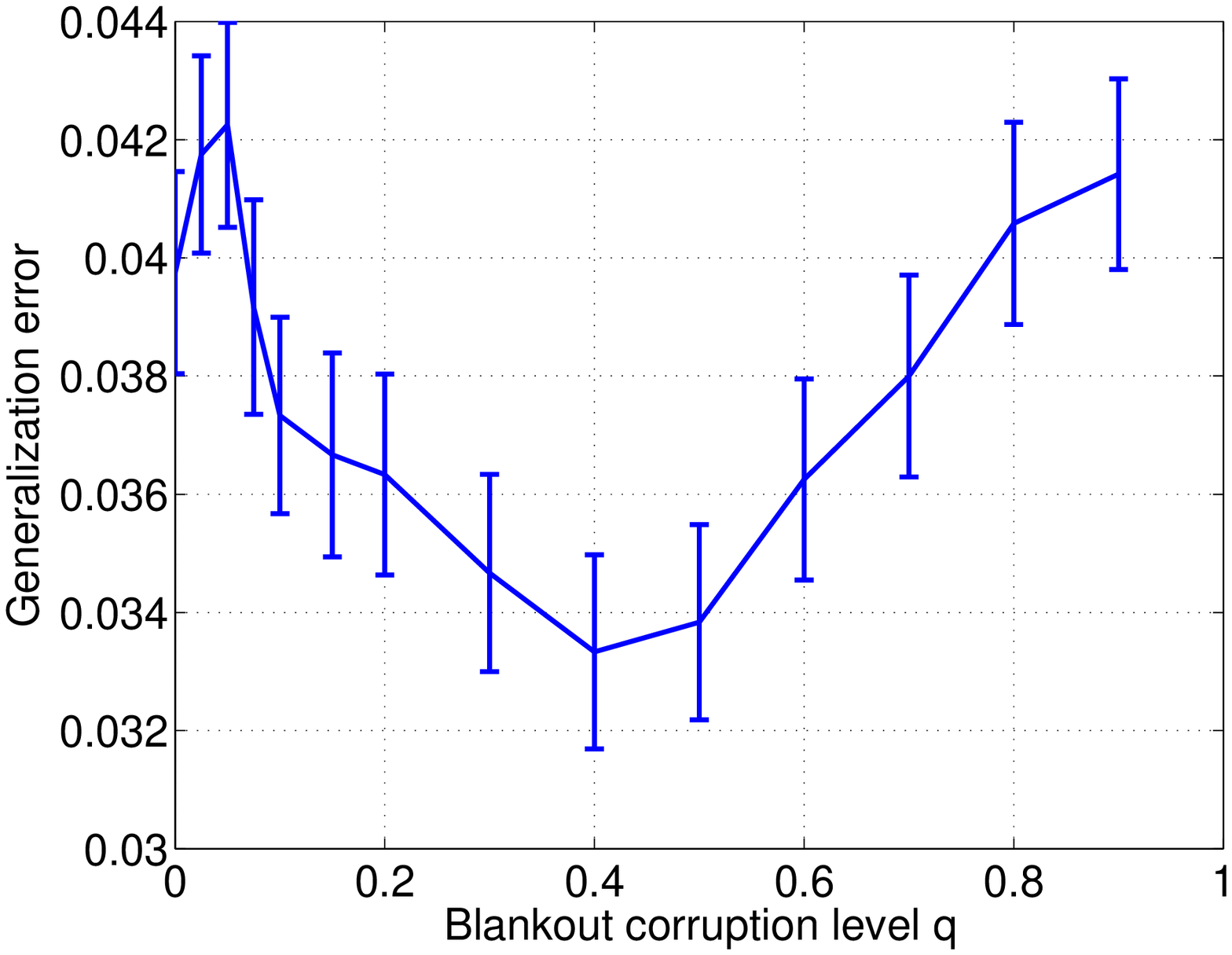}\hspace{.03\textwidth}\includegraphics[width=.48\textwidth]{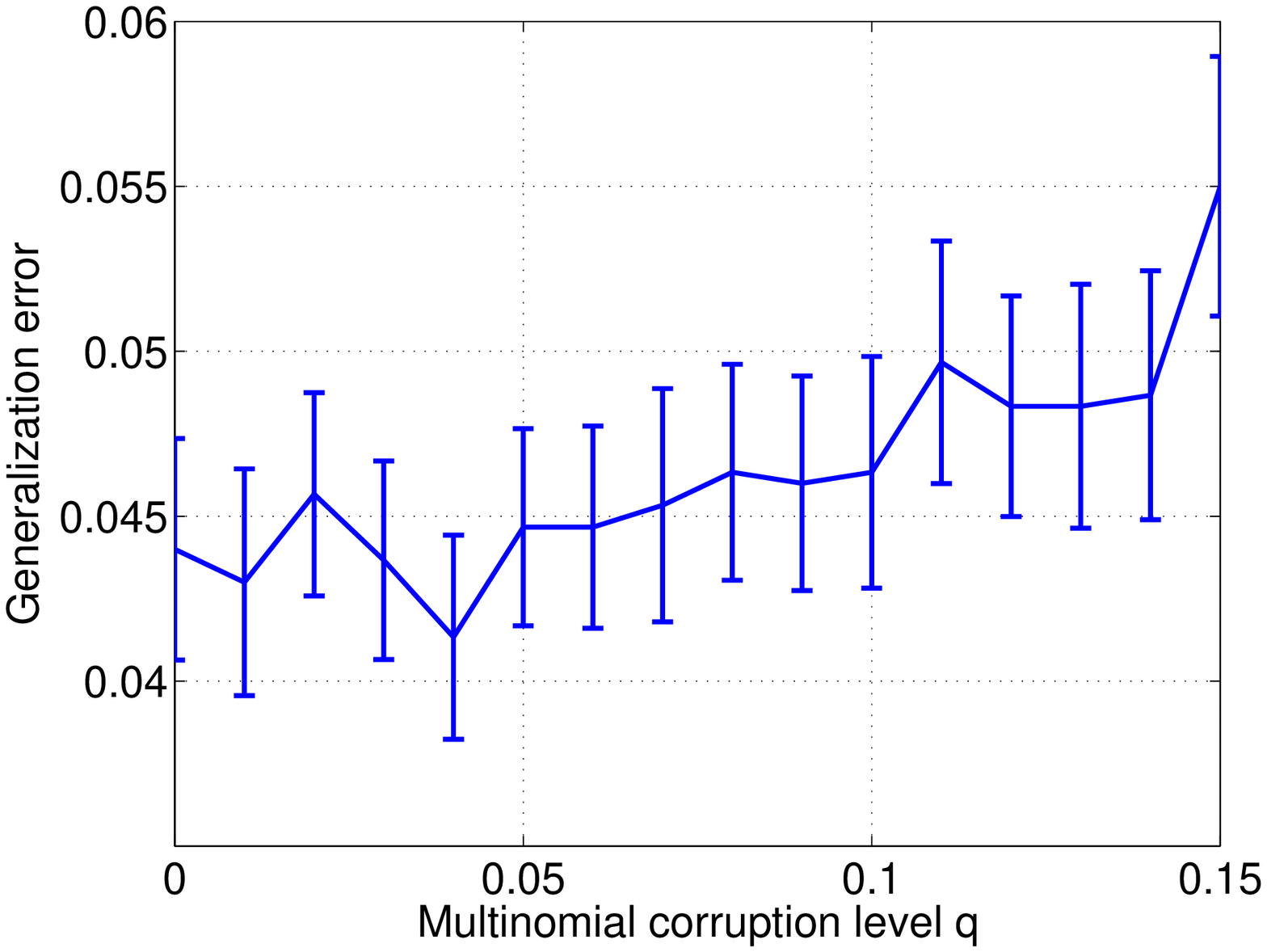}}
\caption{Generalization error of splice-junction classifiers trained using MCF-logistic loss with blankout corruption (left) and multinomial corruption (right) as a function of the corruption rate $q$ (lower is better). All results are obtained by averaging over $50$ runs of $10$-fold cross-validation. The error bars indicate the standard error of the generalization error estimate.}
\label{fig:splice}
\end{figure}

\paragraph{Results.} Figure~\ref{fig:splice} presents the performance of our MCF classifiers with blankout corruption (left) and multinomial corruption (right) as a function of the MCF corruption rate $q$. The results presented in the figure show that MCF with blankout may improve the performance of linear models for the recognition of splice-junctions by more than $0.6\%$, which corresponds to a relative improvement of approximately $15\%$. Improvements in performance are obtained for a wide range of blankout rates, with optimal blankout rates ranging between $0.15$ and $0.4$. The results also show that the multinomial corruption model is too simple to lead to performance improvements. However, we expect that these results can be substantially improved by incorporating in the corruption process knowledge about: 1. the \emph{coverage} obtained on a particular base pair (base-pair reads with higher coverage are read in more DNA snippets and are, therefore, less likely to be erroneous) and 2. knowledge on the error distribution of a DNA sequencing machine (for instance, sequencing machines are more likely to confuse A and T than A and G). Both types of knowledge are available for modern DNA sequencing processes \citep{le13,zagordi10}. The MCF framework allows us to model such domain knowledge into the the corruption model. However, since no information on the coverage and sequencing error distribution is available for our data set \citep{noordewier91}, we leave such extensions of the DNA experiments to future work.

\subsection{Nightmare at test time}
\label{Nightmare at test time}
To test the performance of our MCF predictors with blankout corruption under the ``nightmare at test time'' scenario, we perform experiments on the MNIST data set of handwritten digit images. The MNIST data set contains $N = 60,000$ training and $10,000$ test images of size $D = 28  \times  28  =  784$ pixels, and comprises $K = 10$ classes. 


\begin{figure}[t]
\centering
\centerline{\includegraphics[width=0.8\textwidth]{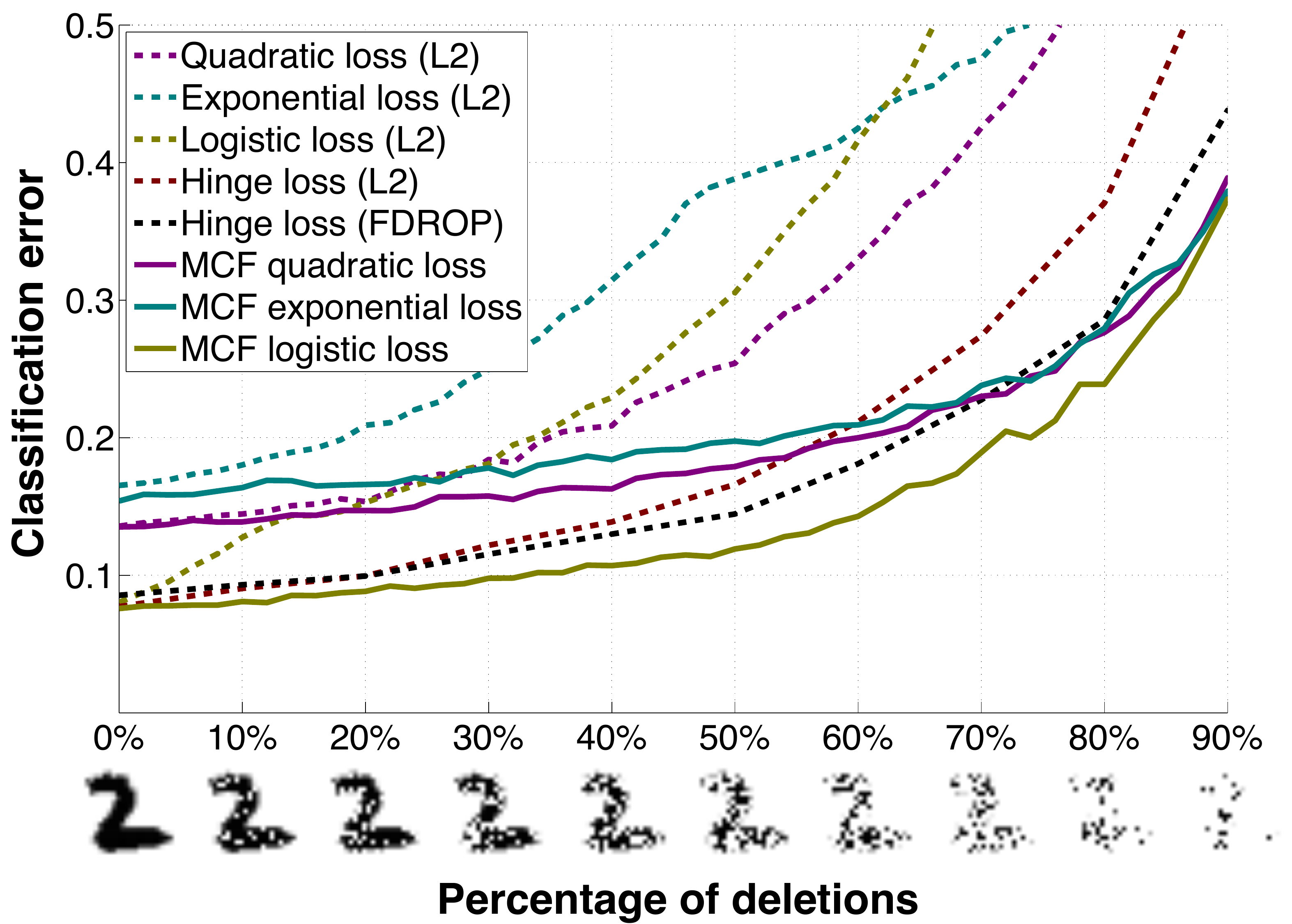}}
\caption{Evaluation on the ``nightmare at test-time'' scenario. Classification errors of standard and MCF predictors with a blankout corruption model  -- trained using three different losses -- and of FDROP \citep{globerson06} on the MNIST data set using the ``nightmare at test time'' scenario. Classification errors are represented on the $y$-axis, whereas the amount of features that are deleted out at test time is represented on the $x$-axis. The images of the digit illustrate the amount of feature deletions applied on the digit images that are used as test data. Figure best viewed in color.}
\label{mnist_results}
\end{figure}
\paragraph{Setup.} We train our predictors on the full training set, and evaluate their performance on versions of the test set in which a certain percentage of the pixels are randomly blanked out, \emph{i.e.} set to zero. We compare the performance of our MCF-predictors (using blankout corruption) with that of standard predictors that use $l_1$ or $l_2$-regularized quadratic, exponential, logistic, and hinge loss. As before, we used the Jensen's upper bound for predictors with MCF logistic loss, and we use cross-validation to determine the optimal value of the regularization parameter. For MCF predictors, we also cross-validate over the blankout corruption level $q$ (again, we use the same noise level for each feature, \emph{i.e.} $\forall d: q_d = q$). In addition to the comparisons with standard predictors, we also compare the performance of MCF with that of FDROP \citep{globerson06}, which is a state-of-the-art algorithm for the ``nightmare at test time'' setting that minimizes the hinge loss under an adversarial worst-case scenario (see section~\ref{Related work}).

The performances are reported as a function of the feature-deletion percentage in the test set, \emph{i.e.} as a function of the probability with which a pixel in the test set is switched off. Following the experimental setting of \citet{globerson06}, we perform the cross-validation for each deletion percentage independently, \emph{i.e.} we create a small validation set with the same feature-deletion level and use it to determine the best regularization parameters and blankout corruption level $q$ for that percentage of feature deletions. We are thus assuming a transductive learning setting, which is a common assumption in many learning scenarios with domain shift \citep{blitzer2007biographies, daume07}.

\paragraph{Results.} Figure~\ref{mnist_results} shows the performance of our predictors as a function of the percentage of deletions in the test images. The figure 
shows the performance for all three loss functions with MCF (solid lines) and without MCF (dashed lines). The performance of a standard predictor using hinge loss is shown as a red dashed line; the performance of FDROP (a state-of-the-art method for this learning setting) is shown as a black dashed line. 
The results presented in Figure~\ref{mnist_results} clearly illustrate the ability of MCF with blankout corruption to produce predictors that are more robust to the ``nightmare at test time'' scenario: MCF improves the performance substantially for all three loss functions considered. For instance, in the case in which $50\%$ of the pixels in the test images is deleted, the performance improvements obtained using MCF for quadratic, exponential, and logistic loss are $40\%$, $47\%$, and $60\%$, respectively. 
Further, the results also indicate that MCF-losses may outperform FDROP; in particular, our average-case logistic loss outperforms FDROP's worst-case hinge loss across the board\footnote{Quadratic and exponential losses perform somewhat worse because they are less appropriate for linear classifiers, but even they outperform FDROP for large numbers of feature deletions in the test data.}. This is particularly impressive as FDROP uses the hinge loss, which performs surprisingly better than the alternative losses on this data set (the improvement of FDROP over the generic hinge-loss is in fact relatively modest). This result suggests that it is better to consider an average-case scenario than a worst-case scenario in the ``nightmare at test time'' setting.

\section{Discussion and Conclusion}
\label{Discussion}

We presented an approach to learn classifiers by marginalizing corrupted features (MCF). Specifically, MCF trains predictors by introducing corruption on the training examples, which is marginalized out in the expectation of the loss function. 
We minimize the expected loss with respect to the model parameters. Our experimental results show that MCF predictors with blankout and Poisson corruption perform very well in the context of bag-of-words features, whilst MCF with blankout corruption was also very effective on DNA data. MCF with blankout corruption appears to be most effective in high-dimensional learning problems in which the data is sparse. MCF with Poisson corruption is particularly interesting for count features, as it improves classification performances without introducing any additional hyper-parameters. As a disclaimer, care must be taken when applying MCF with Poisson corruption on data sets with outliers. The Poisson corruption may strongly emphasize outliers in the expected loss because the variance of a Poisson distribution is identical to its mean, and because we use loss functions that are not robust to outliers. A simple solution to this problem may be to redefine the corruption distribution to $p(\noisyx_d | x_d) = Pois(\noisyx_{nd} | \min\{x_{nd}, u\})$ for some cutoff parameter $u \geq 0$.

Throughout our experiments with MCF on exponential and logistic losses, the weight of the $l_2$-regularization term, which was set by cross-validation, typically ended up  very close to zero. This implies that the regularizing effect of MCF corruption is sufficient even for high dimensional data. In comparison with traditional regularization, MCF corruption shines in two ways: 1. it often yields superior classification performance and 2. it can be much more intuitive to set parameters about data corruption than about model hyper parameters. 
Further, MCF with blankout corruption also appears to prevent \emph{weight undertraining} \citep{sutton05}: it encourages the weight on each feature to be non-zero, in case this particular feature survives the corruption. 

The resulting redundancy is one of the key factors why MCF corruption makes classifiers so effective against the ``nightmare at test-time'' scenario. 
Consequently, MCF classifiers are particularly useful in learning settings in which features in the test data may be missing (\emph{e.g.}, because sensors that were measuring these features temporarily broke down). Learning with MCF is quite different from previous approaches for this setting \citep{dekel08,globerson06}. In particular, prior work focuses on the worst-case scenario whereas we explicitly consider the (arguably) more common average-case scenario by considering all possible corrupted observations. This has the advantage that it is computationally much cheaper and that it allows for incorporating prior knowledge in the learning. For instance, if the data is generated by a collection of unreliable sensors, knowledge on the reliability of a sensor may be used to set the corresponding $q_d$-parameter. The strong performance of MCF in the ``nightmare at test-time'' scenario suggest it be well suited for learning under domain shift \citep{blitzer2007biographies, daume07}. In particular, as the corrupting distribution may be used to shift the data distribution in the source domain towards the data distribution in the target domain --- potentially, by learning the parameters of the corrupting distribution using maximum likelihood. We leave such investigations for future work.

Another interesting direction for future work is to investigate extensions of MCF to structured prediction \citep{wang13b}, as well as to investigate if MCF can be employed for kernel machines. We also plan to explore in more detail what corruption models $p(\noisybx | \bx)$ are useful for what types of data. Further, MCF could be used in the training of neural networks with a single layer of hidden units: blankout noise \emph{on the hidden nodes} can improve the performance of the networks \citep{lecun90,hinton12} and can be marginalized out analytically. Moreover, classifiers trained using MCF-blankout may be very well suited for application in \emph{anytime classification} scenarios \citep{grubb12}, as they are optimized to work well on all possible subsets of features. Another interesting direction for future work is to investigate the theoretical properties of learning with MCF. \cite{xu09} have derived generalization bounds for learning with corruption under the worst-case scenario (\emph{i.e.} for the related models discussed in Section~\ref{Related work}), and \cite{mcallester13} recently derived a PAC-Bayesian bound for learning linear models with blankout in the average-case scenario; however, generic bounds for learning with MCF do not yet exist. A final interesting direction would be to investigate the effect of marginalizing \emph{corrupted labels} or target values \citep{lawrence01,chen13}.

% ...

\section*{Acknowledgements}
The authors thank Fei Sha, Wouter Kouw, Max Welling, and Lawrence Saul for helpful discussions. In particular, the variational upper bound for the logistic loss was suggested by Lawrence Saul. Laurens van der Maaten acknowledges support from the Netherlands Organisation for Scientific Research (NWO; grant 612.001.301), from the EU-FP7 SSPNet and INSIDDE projects, and from the AAL project SALIG++. Kilian Weinberger, Minmin Chen, and Stephen Tyree are supported by NSF grants 1149882 and 1137211.

\appendix
\section{Overview of loss functions}\label{Overview of loss functions}
Table~\ref{table:overview3} gives an overview of all MCF loss functions discussed in the paper. Table~\ref{table:overview4} lists the associated gradients required for learning. For an overview of moment-generating functions and their gradients, we refer to Table~\ref{table:overview2}.

\begin{table}[t]
\begin{center}
\begin{tabular}{|l|l|l|}\hline
\bf MCF Loss & \bf $\by$ & \bf Loss function\\\hline
Quadratic & $\{-1, +1\}$ & $\bw^\top \bigg(\mathbb{E}[\noisybx_n] \mathbb{E}[\noisybx_n]^\top + \mathbb{V}[\noisybx_n] \bigg) \bw - 2\bigg(y_n \mathbb{E}\left[\noisybx_n \right]\bigg)^\top \bw + 1$\\
\rowcolor{gray} Exponential & $\{0, 1\}$ & $\prod_{d=1}^D \mathbb{E}\left[\exp\left(-y_n w_d \noisyx_{nd}\right)\right]$\\
Log.--Quadr.& $\{-1, +1\}$ & $-y\bw^\top\bx + A(\bw^\top \bx) + \frac{1}{2}\frac{\partial^2 A(\bw^\top \bx)}{\partial (\bw^\top \bx)^2}\mathbb{V}[\bw^\top \noisybx_n]$\\
\rowcolor{gray} Log.--Jensen & $\{0, 1\}$ & $\log\bigg(1 + \prod_{d=1}^D \mathbb{E}\left[\exp\left(-y_n w_d \noisyx_{nd}\right)\right]\bigg)$ \\
Log.--Variat. & $\{0, 1\}$ & \parbox{10cm}{$\min_{\lambda \in [0, 1]} \Bigg[\log\bigg(\prod_{d=1}^D\mathbb{E}[(\exp(\lambda y_n w_d \noisyx_{nd})] + $\\$~~~~~~~~~~~~~~~~~~\prod_{d=1}^D \mathbb{E}[\exp((\lambda - 1) y_n w_d \noisyx_{nd})]\bigg)  - \lambda y_n \bw^\top \bx_n\Bigg]$}\\\hline
\end{tabular}
\caption{MCF losses on a labeled instance $(\bx_n, y_n)$. For an overview of moment-generating functions, please refer to Table~\ref{table:overview2}. All the expectations and variances are under the corrupting distribution $p(\noisybx_n | \bx_n)$.}\label{table:overview3}
\end{center}
\end{table}

\begin{savenotes}
\begin{table}[t]
\begin{center}
\resizebox{1.0\textwidth}{!}{
\begin{tabular}{|l|l|}\hline
\bf MCF Loss & \bf Loss gradient\\\hline
Quadratic & Not applicable\footnote{For MCF quadratic loss, the optimal parameters can be identified via the closed-form solution in (\ref{eq:optimal_w_lls}).} \\
\rowcolor{gray} Exponential & $\frac{\partial \mathbb{E}\left[\exp\left(-y_n w_d \noisyx_{nd}\right)\right]}{\partial w_d} \prod_{d' \neq d} \mathbb{E}\left[\exp\left(-y_n w_{d'} \noisyx_{nd'}\right)\right]$\\
Log.--Quadr.& $-y_n\bx_n + 2\sigma(\bw^\top \bx_n) - 1 + \frac{\partial^2 A(\bw^\top \bx_n)}{\partial (\bw^\top \bx_n)^2} (\mathbb{E}[\noisybx_n \noisybx_n^\top] - \bx_n\bx_n^\top) \bw + \frac{1}{2} \frac{\partial^3 A(\bw^\top \bx_n)}{\partial (\bw^\top \bx_n)^3} \mathbb{V}[\bw^\top \noisybx_n]$\\
\rowcolor{gray} Log.--Jensen & $\frac{\prod_{d' \neq d} \mathbb{E}\left[\exp\left(-y_n w_{d'} \noisyx_{nd'}\right)\right]}{1 + \prod_{d'=1}^D \mathbb{E}\left[\exp\left(-y_n w_{d'}\noisyx_{nd'}\right)\right]}\frac{\partial \mathbb{E}\left[\exp\left(-y_n w_d \noisyx_{nd}\right)\right]}{\partial w_d}$ \\
Log.--Variat. & 
\parbox{12cm}{$ -\lambda^* y_n x_{nd} + \frac{\exp(\lambda^* ) \prod_{d' \neq d} \mathbb{E}\left[\exp\left(-y_n w_{d'} \noisyx_{nd'}\right)\right]}
{\prod_{d'=1}^D \mathbb{E}\left[\exp\left(\lambda^* y_n w_{d'} \noisyx_{nd'}\right)\right] + \prod_{d'=1}^D \mathbb{E}\left[\exp\left((\lambda^* - 1) y_n w_{d'} \noisyx_{nd'}\right)\right]}
\frac{\partial \mathbb{E}\left[\exp\left(-y_n w_d \noisyx_{nd}\right)\right]}{\partial w_d}$ \\
$\left[\prod_{d'=1}^D \mathbb{E}\left[\exp\left((\lambda^* - 1)y_n w_{d'} \noisyx_{nd'}\right)\right] + \exp(-1) \prod_{d'=1}^D \mathbb{E}\left[\exp\left((\lambda^* - 2)y_n w_{d'} \noisyx_{nd'}\right)\right] \right]$}\\\hline
\end{tabular}
}
\caption{Derivatives required for learning from a labeled instance $(\bx_n, y_n)$ using various MCF losses. For an overview of the derivatives of moment-generating functions, please refer to Table~\ref{table:overview2}. All the expectations are under the corrupting distribution $p(\noisybx_n | \bx_n)$. In the variational upper bound of the logistic loss, $\lambda^*$ is obtained by performing the minimization described in Table~\ref{table:overview3}.}\label{table:overview4}
\end{center}
\end{table}
\end{savenotes}

\section{Numerical stability}\label{Numerical stability}
When using dropout, the Jensen's upper bound on the expected logistic loss is not numerically stable. In particular, the upper bound $\hat{\mathcal{L}}$ takes the following form for multi-class classification:
\begin{equation}
\hat{\mathcal{L}}(\mathcal{D}; \bw) = -\sum_{n=1}^N \log\left[\frac{\prod_{d=1}^D (q_d + (1 - q_d) \exp(\by_n^\top \bw_d x_{nd}))}{\sum_{{\by'}_n}\prod_{d=1}^D (q_d + (1 - q_d) \exp({\by'}_n^\top \bw_d x_{nd}))} \right].
 \end{equation}
As is common in logistic regression, we can evaluate this expression by expressing the nominator and denominator in the log-domain, performing a log-shift, and exponentiating both sides:
\begin{equation}
\hat{\mathcal{L}}(\mathcal{D}; \bw) = -\sum_{n=1}^N \log\left[\frac{\exp\left(s + \sum_{d=1}^D \log\left[q_d + (1-q_d) \exp(\by_n^\top \bw_d x_{nd})\right]\right)}{\sum_{{\by'}_n} \exp\left(s + \sum_{d=1}^D \log\left[q_d + (1-q_d) \exp({\by'}_n^\top \bw_d x_{nd})\right]\right)} \right],
\end{equation}
for some appropriately chosen log-shift $s$. However, the computation of this quantity may be problematic when $\by_n^\top \bw_d x_{nd} > 0$. We resolve this by noting the following equality:
\begin{equation} 
\log\left[q_d + (1-q_d) \exp(\by_n^\top \bw_d x_{nd})\right] = \by_n^\top \bw_d x_{nd} + \log\left[q_d \exp(-\by_n^\top \bw_d x_{nd}) + 1 - q_d \right].
\end{equation}
Depending on the sign of $\by_n^\top \bw_d x_{nd}$, we compute the left-hand side or the right-hand side of this equation.

\vskip 0.2in
\bibliography{paper}

\begin{thebibliography}{52}
\providecommand{\natexlab}[1]{#1}
\providecommand{\url}[1]{\texttt{#1}}
\expandafter\ifx\csname urlstyle\endcsname\relax
  \providecommand{\doi}[1]{doi: #1}\else
  \providecommand{\doi}{doi: \begingroup \urlstyle{rm}\Url}\fi

\bibitem[Abernethy et~al.(2008)Abernethy, Hazan, and Rakhlin]{abernethy08}
J.~Abernethy, E.~Hazan, and A.~Rakhlin.
\newblock Competing in the dark: An efficient algorithm for bandit linear
  optimization.
\newblock In \emph{Proceedings of the Conference on Learning Theory}, pages
  263--274, 2008.

\bibitem[Allen(1974)]{allen74}
D.M. Allen.
\newblock The relationship between variable selection and data augmentation and
  a method for prediction.
\newblock \emph{Technometrics}, 16:\penalty0 125--127, 1974.

\bibitem[Bhattacharyya et~al.(2004)Bhattacharyya, Pannagadatta, and
  Smola]{bhattacharyya04}
C.~Bhattacharyya, K.S. Pannagadatta, and A.J. Smola.
\newblock A second order cone programming formulation for classifying missing
  data.
\newblock In \emph{Advances in Neural Information Processing Systems}, pages
  153--160, 2004.

\bibitem[Blitzer et~al.(2007)Blitzer, Dredze, and
  Pereira]{blitzer2007biographies}
J.~Blitzer, M.~Dredze, and F.~Pereira.
\newblock Biographies, {B}ollywood, boom-boxes and blenders: Domain adaptation
  for sentiment classification.
\newblock In \emph{Association for Computational Linguistics}, volume~45, page
  440, 2007.

\bibitem[Boyd and Vandenberghe(2004)]{boyd04}
S.P. Boyd and L.~Vandenberghe.
\newblock \emph{Convex optimization}.
\newblock Cambridge University Press, New York, NY, 2004.

\bibitem[Burges and Sch{\"o}lkopf(1997)]{burges1997improving}
C.J.C. Burges and B.~Sch{\"o}lkopf.
\newblock Improving the accuracy and speed of support vector machines.
\newblock \emph{Advances in Neural Information Processing Systems}, 9:\penalty0
  375--381, 1997.

\bibitem[Cesa-Bianchi et~al.(2010)Cesa-Bianchi, Shalev-Shwartz, and
  Shamir]{cesabianchi10}
N.~Cesa-Bianchi, S.~Shalev-Shwartz, and O.~Shamir.
\newblock Online learning of noisy data with kernels.
\newblock In \emph{Proceedings of the Conference on Learning Theory}, pages
  218--230, 2010.

\bibitem[Cesa-Bianchi et~al.(2011)Cesa-Bianchi, Shalev-Shwartz, and
  Shamir]{cesabianchi11}
N.~Cesa-Bianchi, S.~Shalev-Shwartz, and O.~Shamir.
\newblock Online learning of noisy data.
\newblock \emph{IEEE Transactions on Information Theory}, 57\penalty0
  (12):\penalty0 7907--7931, 2011.

\bibitem[Chapelle et~al.(2000)Chapelle, Weston, Bottou, and Vapnik]{chapelle00}
O.~Chapelle, J.~Weston, L.~Bottou, and V.~Vapnik.
\newblock Vicinal risk minimization.
\newblock In \emph{Advances in Neural Information Processing Systems}, pages
  416--422, 2000.

\bibitem[Chechik et~al.(2008)Chechik, Heitz, Elidan, Abbeel, and
  Koller]{chechik08}
G.~Chechik, G.~Heitz, G.~Elidan, P.~Abbeel, and D.~Koller.
\newblock Max-margin classification of data with absent features.
\newblock \emph{Journal of Machine Learning Research}, 9\penalty0
  (Jan):\penalty0 1--21, 2008.

\bibitem[Chen et~al.(2012)Chen, Xu, Weinberger, and Sha]{chen2012msda}
M.~Chen, Z.~Xu, K.Q. Weinberger, and F.~Sha.
\newblock Marginalized denoising autoencoders for domain adaptation.
\newblock In \emph{Proceedings of the International Conference on Machine
  Learning}, pages 767--774, 2012.

\bibitem[Chen et~al.(2013)Chen, Zheng, and Weinberger]{chen13}
M.~Chen, A.~Zheng, and K.Q. Weinberger.
\newblock Fast image tagging.
\newblock In \emph{Proceedings of $30^{th}$ International Conference on Machine
  Learning}, pages 1274--1282, 2013.

\bibitem[Coates et~al.(2011)Coates, Lee, and Ng]{coates11}
A.~Coates, H.~Lee, and A.Y. Ng.
\newblock An analysis of single-layer networks in unsupervised feature
  learning.
\newblock In \emph{Proceedings of the International Conference on Artificial
  Intelligence \& Statistics, JMLR W\&CP 15}, pages 215--223, 2011.

\bibitem[Cover and Hart(1967)]{cover1967nearest}
T.~Cover and P.~Hart.
\newblock Nearest neighbor pattern classification.
\newblock \emph{IEEE Transactions on Information Theory}, 13\penalty0
  (1):\penalty0 21--27, 1967.

\bibitem[Daum{\'e~III}(2007)]{daume07}
H.~Daum{\'e~III}.
\newblock Frustratingly easy domain adaptation.
\newblock In \emph{Proceedings of the $45^{th}$ Annual Meeting of the
  Association of Computational Linguistics}, pages 256--263, 2007.

\bibitem[Dekel and Shamir(2008)]{dekel08}
O.~Dekel and O.~Shamir.
\newblock Learning to classify with missing and corrupted features.
\newblock In \emph{Proceedings of the International Conference on Machine
  Learning}, pages 216--223, 2008.

\bibitem[Duda et~al.(2001)Duda, Hart, and Stork]{duda01}
R.O. Duda, P.E. Hart, and D.G. Stork.
\newblock \emph{Pattern Classification}.
\newblock Wiley Interscience Inc., 2001.

\bibitem[Flaxman et~al.(2005)Flaxman, Kalai, and McMahan]{flaxman05}
A.~Flaxman, A.~Kalai, and H.~McMahan.
\newblock Online convex optimization in the bandit setting: Gradient descent
  without a gradient.
\newblock In \emph{Proceedings of the ACM-SIAM Symposium on Discrete
  Algorithms}, pages 385--394, 2005.

\bibitem[Freund and Schapire(1995)]{freund95}
Y.~Freund and R.E. Schapire.
\newblock A decision-theoretic generalization of on-line learning and an
  application to boosting.
\newblock In \emph{Proceedings of the Second European Conference on
  Computational Learning Theory}, pages 23--37, 1995.

\bibitem[Globerson and Roweis(2006)]{globerson06}
A.~Globerson and S.~Roweis.
\newblock Nightmare at test time: Robust learning by feature deletion.
\newblock In \emph{Proceedings of the International Conference on Machine
  Learning}, pages 353--360, 2006.

\bibitem[Glorot et~al.(2011)Glorot, Bordes, and Bengio]{glorot11}
X.~Glorot, A.~Bordes, and Y.~Bengio.
\newblock Domain adaptation for large-scale sentiment classification: A deep
  learning approach.
\newblock In \emph{Proceedings of the International Conference on Machine
  Learning}, pages 513--520, 2011.

\bibitem[Grave et~al.(2011)Grave, Obozinski, and Bach]{grave11}
E.~Grave, G.~Obozinski, and F.~Bach.
\newblock Trace lasso: {A} trace norm regularization for correlated designs.
\newblock In \emph{Advances in Neural Information Processing Systems 24}, pages
  2187--2195, 2011.

\bibitem[Grubb and Bagnell(2012)]{grubb12}
A.~Grubb and J.A. Bagnell.
\newblock Speedboost: Anytime prediction with uniform near-optimality.
\newblock In \emph{Proceedings of the International Conference on Artificial
  Intelligence \& Statistics (AI-STATS)}, pages 458--466, 2012.

\bibitem[Herbrich and Graepel(2004)]{herbrich2004invariant}
R.~Herbrich and T.~Graepel.
\newblock Invariant pattern recognition by semidefinite programming machines.
\newblock In \emph{Advances in Neural Information Processing Systems},
  volume~16, page~33, 2004.

\bibitem[Hinton et~al.(2012)Hinton, Srivastava, Krizhevsky, Sutskever, and
  Salakhutdinov]{hinton12}
G.E. Hinton, N.~Srivastava, A.~Krizhevsky, I.~Sutskever, and R.R.
  Salakhutdinov.
\newblock Improving neural networks by preventing co-adaptation of feature
  detectors, 2012.

\bibitem[Krizhevsky(2009)]{krizhevsky09}
A.~Krizhevsky.
\newblock Learning multiple layers of features from tiny images.
\newblock Technical report, University of Toronto, 2009.

\bibitem[Lawrence and Sch{\"o}lkopf(2001)]{lawrence01}
N.D. Lawrence and B.~Sch{\"o}lkopf.
\newblock Estimating a kernel {F}isher discriminant in the presence of label
  noise.
\newblock In \emph{Proceedings of the International Conference in Machine
  Learning}, pages 306--313, 2001.

\bibitem[Le et~al.(2013)Le, Schulz, McCauley, Hinman, and Bar-Joseph]{le13}
H.S. Le, M.H. Schulz, B.M. McCauley, V.F. Hinman, and Z.~Bar-Joseph.
\newblock Probabilistic error correction for {RNA} sequencing.
\newblock \emph{Nucleic Acids Research}, 41\penalty0 (10):\penalty0 e109, 2013.

\bibitem[LeCun et~al.(1990)LeCun, Denker, and Solla]{lecun90}
Y.~LeCun, J.S. Denker, and S.A. Solla.
\newblock Optimal brain damage.
\newblock In \emph{Advances in Neural Information Processing Systems}, pages
  598--605, 1990.

\bibitem[Livni et~al.(2012)Livni, Crammer, and Globerson]{livni12}
R.~Livni, K.~Crammer, and A.~Globerson.
\newblock A simple geometric interpretation of svm using stochastic
  adversaries.
\newblock In \emph{Proceedings of the $15^{th}$ International Conference on
  Artificial Intelligence and Statistics (AI-STATS)}, pages 722--730, 2012.

\bibitem[Maillet et~al.(2009)Maillet, Eck, Desjardins, and Lamere]{maillet09}
F.~Maillet, D.~Eck, G.~Desjardins, and P.~Lamere.
\newblock Steerable playlist generation by learning song similarity from radio
  station playlists.
\newblock In \emph{Proceedings of the International Society for Music
  Information Retrieval Conference}, pages 345--350, 2009.

\bibitem[McAllester(2013)]{mcallester13}
D.~McAllester.
\newblock A {PAC}-{B}ayesian tutorial with a dropout bound, 2013.

\bibitem[Mesnil et~al.(2011)Mesnil, Dauphin, Glorot, Rifai, Bengio, Goodfellow,
  Lavoie, Muller, Desjardins, Warde-Farley, Vincent, Courville, and
  Bergstra]{mesnil11}
G.~Mesnil, Y.~Dauphin, X.~Glorot, S.~Rifai, Y.~Bengio, I.~Goodfellow,
  E.~Lavoie, X.~Muller, G.~Desjardins, D.~Warde-Farley, P.~Vincent,
  A.~Courville, and J.~Bergstra.
\newblock Unsupervised and transfer learning challenge: a deep learning
  approach.
\newblock \emph{JMLR: Workshop and Conference Proceedings}, 7:\penalty0 1--15,
  2011.

\bibitem[Noordewier et~al.(1991)Noordewier, Towell, and Shavlik]{noordewier91}
M.O. Noordewier, G.G. Towell, and J.W. Shavlik.
\newblock Training knowledge-based neural networks to recognize genes in {DNA}
  sequences.
\newblock In \emph{Advances in Neural Information Processing Systems},
  volume~3, pages 530--536, 1991.

\bibitem[Parzen(1962)]{parzen62}
E.~Parzen.
\newblock On estimation of a probability density function and mode.
\newblock \emph{Annals of Mathematical Statistics}, 33:\penalty0 1065--1076,
  1962.

\bibitem[Ranzato and Hinton(2010)]{ranzato10}
M.~Ranzato and G.E. Hinton.
\newblock Modeling pixel means and covariances using factorized third-order
  {B}oltzmann machines.
\newblock In \emph{Proceedings of the IEEE Conference on Computer Vision and
  Pattern Recognition}, pages 2551--2558, 2010.

\bibitem[Rostamizadeh et~al.(2011)Rostamizadeh, Agarwal, and
  Bartlett]{rostamizadeh11}
A.~Rostamizadeh, A.~Agarwal, and P.~Bartlett.
\newblock Learning with missing features.
\newblock In \emph{Proceedings of Uncertainty in Artificial Intelligence},
  pages 635--642, 2011.

\bibitem[Sha and Pereira(2003)]{sha03}
F.~Sha and F.~Pereira.
\newblock Shallow parsing with {C}onditional {R}andom {F}ields.
\newblock In \emph{Proceedings of Human Language Technology -- NAACL 2003},
  pages 213--220, 2003.

\bibitem[Shafer and Vovk(2008)]{shafer2008tutorial}
G.~Shafer and V.~Vovk.
\newblock A tutorial on conformal prediction.
\newblock \emph{The Journal of Machine Learning Research}, 9:\penalty0
  371--421, 2008.

\bibitem[Shivaswamy et~al.(2006)Shivaswamy, Bhattacharyya, and
  Smola]{shivaswamy06}
P.K. Shivaswamy, C.~Bhattacharyya, and A.J. Smola.
\newblock Second order cone programming approaches for handling missing and
  uncertain data.
\newblock \emph{Journal of Machine Learning Research}, 7:\penalty0 1283--1314,
  2006.

\bibitem[Sutton et~al.(2005)Sutton, Sindelar, and McCallum]{sutton05}
C.~Sutton, M.~Sindelar, and A.~McCallum.
\newblock Feature bagging: Preventing weight undertraining in structured
  discriminative learning.
\newblock Technical Report IR-402, University of Massachusetts, 2005.

\bibitem[Teo et~al.(2008)Teo, Globerson, Roweis, and Smola]{teo2008convex}
C.H. Teo, A.~Globerson, S.~Roweis, and A.~Smola.
\newblock Convex learning with invariances.
\newblock \emph{Advances in Neural Information Processing Systems},
  20:\penalty0 1489--1496, 2008.

\bibitem[Torralba et~al.(2008)Torralba, Fergus, and Freeman]{torralba08}
A.~Torralba, R.~Fergus, and W.T. Freeman.
\newblock 80 million tiny images: A large dataset for non-parametric object and
  scene recognition.
\newblock \emph{IEEE Transactions on Pattern Analysis and Machine
  Intelligence}, 30\penalty0 (11):\penalty0 1958--1970, 2008.

\bibitem[Trafalis and Gilbert(2007)]{trafalis07}
T.~Trafalis and R.~Gilbert.
\newblock Robust support vector machines for classification and computational
  issues.
\newblock \emph{Optimization Methods and Software}, 22\penalty0 (1):\penalty0
  187--198, 2007.

\bibitem[van~der Maaten et~al.(2013)van~der Maaten, Chen, Tyree, and
  Weinberger]{vandermaaten13}
L.J.P. van~der Maaten, M.~Chen, S.~Tyree, and K.Q. Weinberger.
\newblock Learning by marginalizing corrupted features.
\newblock In \emph{Proceedings of the International Conference on Machine
  Learning}, pages 410--418, 2013.

\bibitem[Vincent et~al.(2008)Vincent, Larochelle, Bengio, and
  Manzagol]{vincent08}
P.~Vincent, H.~Larochelle, Y.~Bengio, and P.A. Manzagol.
\newblock Extracting and composing robust features with denoising autoencoders.
\newblock In \emph{Proceedings of the International Conference on Machine
  Learning}, pages 1096--1103, 2008.

\bibitem[Vincent et~al.(2010)Vincent, Larochelle, Lajoie, Bengio, and
  Manzagol]{vincent10}
P.~Vincent, H.~Larochelle, I.~Lajoie, Y.~Bengio, and P.-A. Manzagol.
\newblock Stacked denoising autoencoders: Learning useful representations in a
  deep network with a local denoising criterion.
\newblock \emph{Journal of Machine Learning Research}, 11\penalty0
  (Dec):\penalty0 3371--3408, 2010.

\bibitem[Wager et~al.(2013)Wager, Wang, and Liang]{wagerArxiv13}
S.~Wager, S.~Wang, and P.~Liang.
\newblock Dropout training as adaptive regularization.
\newblock In \emph{Advances in Neural Information Processing Systems (NIPS)},
  pages 351--359, 2013.

\bibitem[Wang et~al.(2013)Wang, Wang, Wager, Liang, and Manning]{wang13b}
S.~Wang, M.~Wang, S.~Wager, P.~Liang, and C.D. Manning.
\newblock Feature noising for log-linear structured prediction.
\newblock In \emph{Proceedings of the Conference on Empirical Methods in
  Natural Language Processing}, pages 1170--1179, 2013.

\bibitem[Xu et~al.(2009)Xu, Caramanis, and Mannor]{xu09}
H.~Xu, C.~Caramanis, and S.~Mannor.
\newblock Robustness and regularization of support vector machines.
\newblock \emph{Journal of Machine Learning Research}, 10:\penalty0 1485--1510,
  2009.

\bibitem[Zagordi et~al.(2010)Zagordi, Klein, D{\"a}umer, and
  Beerenwinkel]{zagordi10}
O.~Zagordi, R.~Klein, M.~D{\"a}umer, and N.~Beerenwinkel.
\newblock Error correction of next-generation sequencing data and reliable
  estimation of {HIV} quasispecies.
\newblock \emph{Nucleic Acids Research}, 38\penalty0 (21):\penalty0 7400--7409,
  2010.

\bibitem[Zhu et~al.(2006)Zhu, Rosset, Zou, and Hastie]{zhu06}
J.~Zhu, S.~Rosset, H.~Zou, and T.~Hastie.
\newblock Multi-class {A}da{B}oost.
\newblock Technical Report 430, Department of Statistics, University of
  Michigan, 2006.

\end{thebibliography}

\end{document}